\documentclass[journal]{IEEEtran}
\usepackage{cite}

\ifCLASSINFOpdf
   \usepackage[pdftex]{graphicx}
\else
\fi
\usepackage{amsmath}
\usepackage{amssymb}
\usepackage{xcolor}

\ifCLASSOPTIONcompsoc
 \usepackage[caption=false,font=normalsize,labelfont=sf,textfont=sf]{subfig}
\else
 \usepackage[caption=false,font=footnotesize]{subfig}
\fi
\usepackage[unicode=true,
 bookmarks=true,bookmarksnumbered=true,bookmarksopen=true,bookmarksopenlevel=1,
 breaklinks=false,pdfborder={0 0 0},pdfborderstyle={},backref=false,colorlinks=false]
 {hyperref}

\begin{document}
%
\title{Evaluation of automated driving system safety metrics with logged vehicle trajectory data}
%
%
%
\author{Xintao~Yan, Shuo~Feng,~\IEEEmembership{Member,~IEEE}, David~J.~LeBlanc, Carol~Flannagan, Henry~X.~Liu,~\IEEEmembership{Member,~IEEE}

\thanks{This work was supported by the National Highway Traffic Safety Administration (NHSTA) of the US Department of Transportation (USDOT). \emph{(Corresponding authors: Shuo Feng and Henry X. Liu.)}}%
\thanks{Xintao Yan is with the Department of Civil and Environmental
Engineering, University of Michigan, Ann Arbor, MI 48109 USA, (e-mail: xintaoy@umich.edu).}%
\thanks{Shuo~Feng is with the Department of Automation, Tsinghua University, Beijing
100084, China, (e-mail: fshuo@tsinghua.edu.cn).}
\thanks{David~J.~LeBlanc, and Carol~Flannagan are with the University of Michigan Transportation Research
Institute, Ann Arbor, MI 48109 USA, (e-mail: leblanc@umich.edu; cacf@umich.edu).}%
\thanks{Henry~X.~Liu is with the Department of Civil and Environmental Engineering, MCity, and also University
of Michigan Transportation Research Institute, University of Michigan, Ann Arbor, MI 48109 USA, (e-mail: henryliu@umich.edu).}}

\maketitle

\begin{abstract}
Real-time safety metrics are important for the automated driving system (ADS) to assess the risk of driving situations and to assist the decision-making. Although a number of real-time safety metrics have been proposed in the literature, systematic performance evaluation of these safety metrics has been lacking. As different behavioral assumptions are adopted in different safety metrics, it is difficult to compare the safety metrics and evaluate their performance. To overcome this challenge, in this study, we propose an evaluation framework utilizing logged vehicle trajectory data, in that vehicle trajectories for both subject vehicle (SV) and background vehicles (BVs) are obtained and the prediction errors caused by behavioral assumptions can be eliminated. Specifically, we examine whether the SV is in a collision unavoidable situation at each moment, given all near-future trajectories of BVs. In this way, we level the ground for a fair comparison of different safety metrics, as a good safety metric should always alarm in advance to the collision unavoidable moment. When trajectory data from a large number of trips are available, we can systematically evaluate and compare different metrics' statistical performance. In the case study, three representative real-time safety metrics, including the time-to-collision (TTC) \cite{hayward1972near-TTC}, the PEGASUS Criticality Metric (PCM) \cite{junietz2018criticality-PCM} and the Model Predictive Instantaneous Safety Metric (MPrISM) \cite{weng2020-MPrISM}, are evaluated using a large-scale simulated trajectory dataset. The results demonstrate that the MPrISM achieves the highest recall and the PCM has the best accuracy. The proposed evaluation framework is important for researchers, practitioners, and regulators to characterize different metrics, and to select appropriate metrics for different applications. Moreover, by conducting failure analysis on moments when a safety metric failed, we can identify its potential weaknesses which are valuable for its potential refinements and improvements.
\end{abstract}


%
\IEEEpeerreviewmaketitle

\section{Introduction\label{sec:Sec-1}}
%
%
%
%
\IEEEPARstart{R}{eal}-time safety metrics are important for vehicles with different levels of driving automation \cite{SAE-J3016}. By
supervising real-time driving conditions, it can serve as the safety
guard to warn human drivers, activate the advanced driver assistance
system (ADAS) or disengage the automated mode, and assist in the decision-making
module to achieve safe planning and control of autonomous vehicles (AVs).

The basic idea of real-time safety metrics is to assess the situational
safety of the current moment based on the prediction of future trajectories
of both the subject vehicle (SV) and the background vehicles (BVs). As
all predictions are made based on the information up to the evaluated
moment, different behavior assumptions for the future maneuvers of the SV and BVs will lead to distinct safety metrics. For example, to calculate time-to-collision (TTC) \cite{hayward1972near-TTC}, the author adopts the behavior assumption that both SV and BV will maintain their current velocity and heading. The PEGASUS Criticality Metric (PCM) makes similar behavior assumptions regarding BVs but assumes the SV will take evasive maneuvers \cite{junietz2018criticality-PCM}. The Model Predictive Instantaneous Safety Metric (MPrISM), on the other hand, assumes the BV will take the worst-case behavior to challenge the
SV, and the SV is assumed to take the best response trying to avoid a collision \cite{weng2020-MPrISM}.


In addition to applying specific behavior assumptions, \cite{shalev2017formal-RSS,nister2019safety-SFF,pek2020using-Formal-method} consider different potential behaviors of the SV and BVs in different situations to construct the safety envelope. For example, the Responsibility-Sensitive Safety (RSS) model \cite{shalev2017formal-RSS} defines a series of scenario-specific safe longitudinal and lateral distances based on rule-based behavior assumptions with considerations of responsibility, traffic rules, and visibility, etc. \cite{althoff2009model-Collision-Probability-Markov-chain,schreier2016integrated-Collision-Probability-DBN,feng2021intelligent-NADE} try to capture the probability of different future behaviors and take that into account to evaluate the SV situational safety. For example, a risk-based metric was proposed in \cite{feng2021intelligent-NADE}, which estimates the collision probability considering BVs’ probabilistic future behaviors.
More detailed literature reviews regarding existing safety metrics and their behavioral assumptions can be found in 
\cite{weng2021class,weng2022finite,singh2023diversity,dahl2018collision-Review-TIV,wishart2020driving-Review-Wishart,li2020threat-Review-ITSM,wang2021review-Review-AAP}.

Although a number of safety metrics have been proposed in the past, there are
very limited studies on the evaluation and comparison of real-time safety metrics
for AVs. It is also well-known that human driving behaviors are highly heterogeneous and stochastic,
but the impact of different behavior assumptions on the performance of safety metrics has not been investigated. The key problem is the lack of a comparable base that is objective and fair. Most existing evaluation methods rely on expert knowledge or heuristics to determine the comparable base, which suffers
from subjective biases. For example, Lee and Peng \cite{lee2005evaluation-Evaluation}
proposed an evaluation method to compare the performance of five forward
collision warning metrics. The metric performance is evaluated by
examining whether it can alarm at pre-defined dangerous moments. To distinguish whether a moment is
dangerous or safe, manually defined rules were used. Specifically,
a moment will be classified as dangerous if the SV is slower than
its leading vehicle and its braking is harder than 0.23g. van Nunen et al. \cite{van2017evaluation-Evaluation}
applied a similar heuristic method to evaluate three real-time safety metrics
performance in truck platooning applications. Moreover, for both methods, they
can only be applied to car-following situations, which are not sufficient
to evaluate metrics used for AV-related applications. 

To evaluate the performance of real-time safety metrics, in this paper, we propose a systematic framework that leverages logged vehicle trajectory data. We assume that the SV is equipped with perception and localization sensors that can detect, track, and localize all BVs, and the vehicle trajectories of BVs are available after the trip. In this way, when we play back to determine the situational safety of the SV at any moment during the trip, trajectories of BVs are given and no behavior assumptions are needed. Therefore, we can level the play field to evaluate different safety metrics and make fair comparisons. 



To illustrate our evaluation framework, let us
first consider the change of situational safety before a collision
event. Intuitively, the level of dangerousness will be increasing
before a collision. Tracing back from the collision moment, there must be a period of
time when the collision becomes unavoidable. We define a moment as
collision unavoidable if the SV cannot avoid the collision even with
evasive maneuvers. Similarly, before the collision unavoidable time,
there must be a period of unsafe situations, in that if the current
vehicle states continue, a collision might become unavoidable. Evasive
maneuvers (e.g., hard braking or emergency lane-change) of the SV
or BVs are needed to avoid a collision. It is apparent that an ideal real-time safety metric should alarm in advance to the collision unavoidable moment 
so that the SV can have time to take action to avoid the collision. With an identified collision unavoidable time for a trip with collision, we can evaluate and compare various safety metrics to see if they were able to alarm before the collision unavoidable situation and how much time in advance they can alarm the dangerous situation. 

To identify collision unavoidable moments, we design an algorithm
to compute if there exists an evasive trajectory for the SV at each
moment, given all BVs near-future trajectory information obtained
from logged data. By conducting failure analysis on moments that safety metrics
failed, the effects of its behavior assumptions, model approximations, and hyper-parameter selections
can be analyzed. The evaluation processes can also help metric developers
to tune hyper-parameters and refine model approximations
to balance computational efficiency and metric performance. Given a large number of logged trajectories,
we can calculate the confusion matrix for
each real-time safety metric, which can be used to further construct
statistical measurements such as Receiver Operating Characteristic
(ROC) curve and Precision-Recall (PR) curve. By analyzing these measurements,
the characteristics of different safety metrics can be better understood,
which can help researchers, practitioners, and regulators to compare
different metrics and choose from them. To validate the effectiveness
of the proposed evaluation framework, three representative real-time
safety metrics are evaluated in this paper, including the TTC \cite{hayward1972near-TTC}, the PCM \cite{junietz2018criticality-PCM}, and the MPrISM \cite{weng2020-MPrISM}.

The rest of this paper is organized as follows. In Section \ref{sec:Sec-2},
the proposed evaluation framework and the algorithm of determining
collision unavoidable moments using logged vehicle trajectory data will be
illustrated. The evaluation and comparison results of different metrics
will be discussed in Section \ref{sec:Sec-3}. Finally, Section \ref{sec:Sec-4}
will conclude the paper and discuss future research directions.

\section{Evaluation framework with logged trajectory data \label{sec:Sec-2}}

In this section, we propose a systematic framework for performance evaluation of 
real-time safety metrics, leveraging logged vehicle trajectories.
At each moment during the trip, the real-time safety metric is essentially working
as a classifier that will decide to alarm or not. We can then evaluate the recall and accuracy of those alarms using the collision unavoidable moments identified from logged trajectories. The overall framework is described in  Section \ref{subsec:Real-time-safety-metric-as-a-classification-problem}.
To examine whether the SV is collision unavoidable at
a moment, we design an optimization problem to compute if there exists
an evasive trajectory for the SV, given all near-future logged
trajectory information of BVs, as discussed in Section \ref{subsec:Collision-unavoidable-moment-calculation}.
In Section \ref{subsec:Statistical-performance-analysis}, we will
introduce the statistical measurements used in this study for analyzing
the performance of safety metrics based on a large quantity of trajectory data.

\subsection{Determine ground-truth moments that real-time safety metrics should alarm \label{subsec:Real-time-safety-metric-as-a-classification-problem}}

\begin{figure*}[tp]
\centering
\subfloat[\label{fig:real-time-safety-metric-cal}]{\includegraphics
[width=0.9\textwidth]{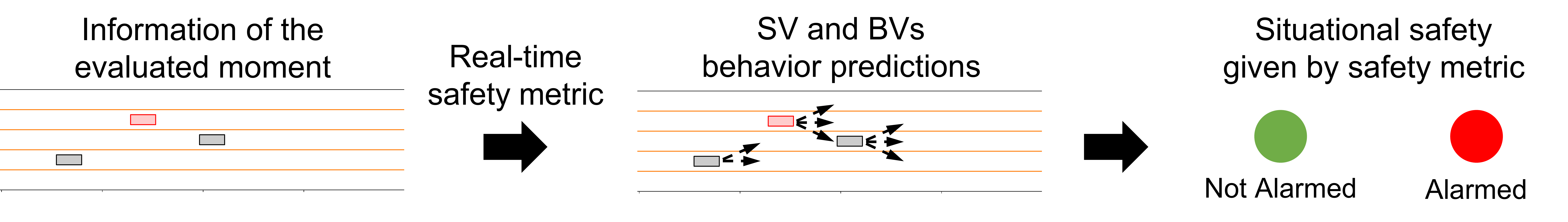}

}

\subfloat[\label{fig:logged-data-cal}]{\includegraphics[width=0.9\textwidth]{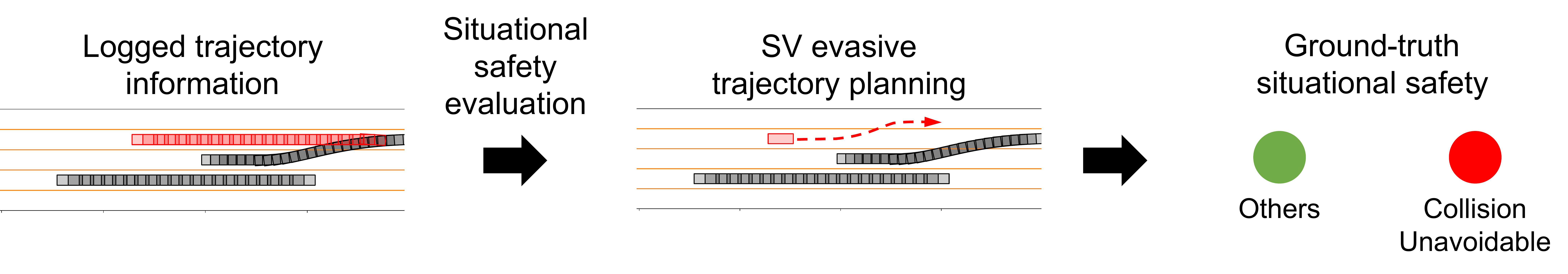}

}

\caption{Comparison of the real-time safety metric calculation (Fig. 1a) and 
the situational safety evaluation using logged vehicle trajectory data (Fig. 1b). The red rectangle
denotes the SV and black rectangles denote BVs.}
\end{figure*}

A real-time safety metric is a mapping $y=\mathcal{F}\left(s\right)$
from the input state information $s$ to the output situational safety
measurement $y$. Different safety metrics may take different inputs
and make different behavior assumptions to calculate the output values
as illustrated in Fig. \ref{fig:real-time-safety-metric-cal}.
For example, the TTC takes the relative distance and relative speed
between the SV and its leading BV as the input $s$ and makes constant
behavior assumptions to calculate the TTC. The metric should detect
dangerous moments and alarm the driver or AV decision module to
react to safety-critical situations. Normally, an activation threshold
$y_{0}$ will be chosen to determine whether the safety metric alarmed.
For example, for the TTC metric, a few existing literature \cite{papadoulis2019evaluating-TTC-1.5s,virdi2019safety-TTC-1.5s}
use $y_{0}=1.5s$ as the threshold so the metric is considered as
alarmed if $TTC\leq1.5s$. Therefore, the situational safety assessment
is essentially a classification problem and each real-time safety
metric is a classifier. To evaluate the performance of different metrics,
we need to analyze the metric outputs and compare them with the situational
safety ground truth. 

However, there is no directly available ground truth and it is difficult to objectively determine whether a moment is dangerous or safe in real-world situations. As we discussed in Section \ref{sec:Sec-1}, for a trip with collision, one of the moments that we can determine objectively from the post-trip perspective using logged trajectory data is the collision unavoidable moment. All safety metrics should alarm at or before that moment. For a trip without collision, an ideal safety metric in our evaluation framework should not alarm because it would be a false alarm from a post-trip perspective. One may argue that a safety metric should alarm for near-miss events, but this will require the identification of near-miss events, which is hard to define in an objective manner and therefore it is beyond the scope of this paper. 

To determine the collision unavoidable moment, we examine
whether there exists an evasive trajectory for the SV, given all near-future
trajectories of BVs after this moment. If there is no feasible evasive
trajectory exists, the SV is in a collision unavoidable situation.
The illustration figure is shown in Fig. \ref{fig:logged-data-cal}.
It should be noted that the proposed
collision unavoidable situation is an objective reflection of the
SV situational safety without relying on expert knowledge or heuristically
defined rules. Comparing with situational safety calculated by the real-time safety
metric, the prediction errors of BV behaviors can be eliminated thanks to logged trajectory information.

\subsection{Identification of collision unavoidable moments  \label{subsec:Collision-unavoidable-moment-calculation}}

To identify a collision unavoidable moment, we will
formulate an optimization problem to compute whether there exists
an evasive trajectory for the SV, given all near-future trajectories
of BVs. 
The formulation is inspired by \cite{weng2020-MPrISM} and follow the framework discussed in \cite{weng2021class}.
We assume the behavior of BVs will not
be influenced by the SV actions within a short look-ahead horizon,
so they will follow their trajectories as observed. 

We adopt the nonlinear vehicle kinematics used in \cite{junietz2018criticality-PCM}
and \cite{weng2020-MPrISM} to model SV dynamics. Let $s\in\mathbb{R}^{4}$
denote the SV state and we have
\begin{align}
\dot{s} & =\left[\begin{array}{cccc}
\dot{p} & \dot{q} & \dot{v} & \dot{\phi}\end{array}\right]^{T}\label{eq: SV dynamics original}\\
 & =\left[\begin{array}{cccc}
v\text{cos}\left(\phi\right) & v\text{sin}\left(\phi\right) & a_{x} & \frac{a_{y}}{v}\end{array}\right]^{T},\label{eq: SV dynamics}
\end{align}
where $(p,q)$ denotes the coordinates of the vehicle, $v$ denotes
the velocity, $\phi$ denotes the heading angle, and $u=[a_{x},a_{y}]^{T}$
denotes the longitudinal and lateral accelerations. With the assumption
of small course angle and small change of velocity, we can linearize
Eq. (\ref{eq: SV dynamics}) as
\begin{equation}
\dot{s}\approx\left[\begin{array}{cccc}
v & \tilde{v}\phi & a_{x} & \frac{a_{y}}{\tilde{v}}\end{array}\right],
\end{equation}
where $\tilde{v}$ denotes the initial state speed. Then the SV system
dynamics can be expressed as
\begin{equation}
s\left(t+1\right)=\text{A}s\left(t\right)+\text{B}u\left(t\right),
\end{equation}
\begin{equation}
\begin{array}{cc}
\text{A}=\left[\begin{array}{cccc}
1 & 0 & \Delta & 0\\
0 & 1 & 0 & \tilde{v}\Delta\\
0 & 0 & 1 & 0\\
0 & 0 & 0 & 1
\end{array}\right], & \text{B}=\left[\begin{array}{cc}
\frac{\Delta^{2}}{2} & 0\\
0 & \frac{\Delta^{2}}{2}\\
\Delta & 0\\
0 & \frac{\Delta}{\tilde{v}}
\end{array}\right],\end{array}
\end{equation}
where $t$ denotes the current timestep and $\Delta$ denotes the
time step size. The vehicle admissible control input $u$ is constrained
by the friction circle, i.e., the Kamm\textquoteright s circle. In
this study, we approximate the ellipse Kamm\textquoteright s circle
using 12 linear inequality constraints as in existing studies, for example, \cite{weng2020-MPrISM}.
It can be represented as
\begin{equation}
\text{G}u\leq\text{h},
\end{equation}
\begin{equation}
\text{G}=\left[\begin{array}{cc}
\text{L}_{x}^{min} & \text{L}_{y}^{max}\\
\text{L}_{x}^{min} & -\text{L}_{y}^{max}\\
-\text{L}_{x}^{max} & \text{L}_{y}^{max}\\
-\text{L}_{x}^{max} & -\text{L}_{y}^{max}
\end{array}\right],\text{h}=\left[\begin{array}{c}
\text{sin}\frac{5\pi}{12}\\
...\\
\text{sin}\frac{5\pi}{12}
\end{array}\right],
\end{equation}
\begin{equation}
\text{L}_{x}^{min}=\left[\begin{array}{c}
\frac{1}{\left|a_{x}^{min}\right|}\text{cos}\left(\frac{7\pi}{12}\right)\\
\frac{1}{\left|a_{x}^{min}\right|}\text{cos}\left(\frac{9\pi}{12}\right)\\
\frac{1}{\left|a_{x}^{min}\right|}\text{cos}\left(\frac{11\pi}{12}\right)
\end{array}\right],\text{L}_{x}^{max}=\left[\begin{array}{c}
\frac{1}{\left|a_{x}^{max}\right|}\text{cos}\left(\frac{7\pi}{12}\right)\\
\frac{1}{\left|a_{x}^{max}\right|}\text{cos}\left(\frac{9\pi}{12}\right)\\
\frac{1}{\left|a_{x}^{max}\right|}\text{cos}\left(\frac{11\pi}{12}\right)\hspace{10pt}
\end{array}\right],\label{}
\end{equation}
\begin{equation}
\text{L}_{y}^{max}=\left[\begin{array}{c}
\frac{1}{\left|a_{y}^{max}\right|}\text{cos}\left(\frac{7\pi}{12}\right)\\
\frac{1}{\left|a_{y}^{max}\right|}\text{cos}\left(\frac{9\pi}{12}\right)\\
\frac{1}{\left|a_{y}^{max}\right|}\text{cos}\left(\frac{11\pi}{12}\right)
\end{array}\right].\label{}
\end{equation}
where $a_{x}^{min},a_{x}^{max}$ are maximum longitudinal deceleration
and acceleration. And $a_{y}^{max}$ is the maximum lateral acceleration
and also deceleration since the lateral maneuver capability is symmetric.

\begin{figure}
\begin{centering}
\includegraphics[scale=0.8]{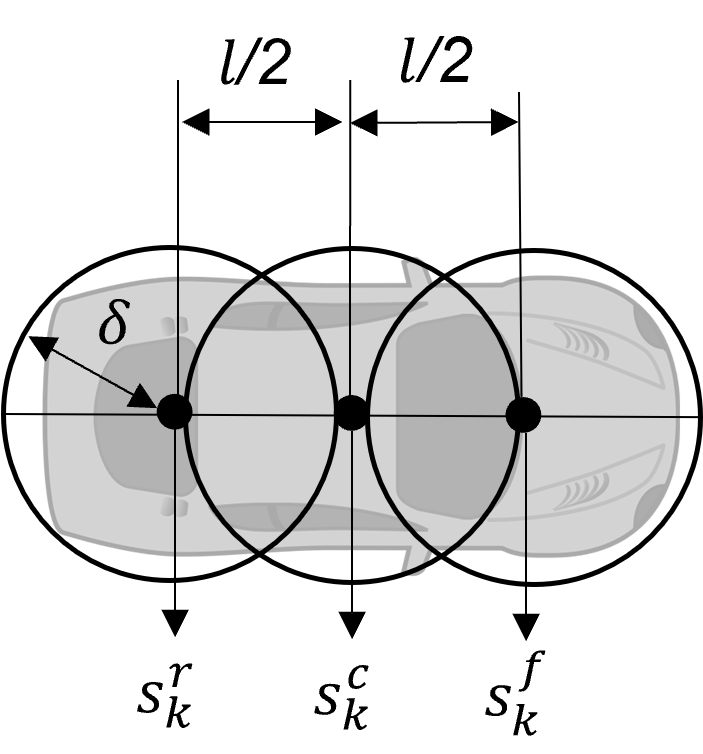}
\par\end{centering}
\caption{The shape of each vehicle is approximated by three circles. \label{fig:3-circles-approx}}
\end{figure}

All vehicles are approximated using three circles with equal radius
$\delta$ as shown in Fig. \ref{fig:3-circles-approx}. The distance
between two consecutive centers is $l$/2. Let $s_{k}^{p},p=\left\{ r,c,f\right\} $
denote the state of each circle center of the $k$-th vehicle. The
rear and front circles\textquoteright{} center $s_{k}^{r},s_{k}^{f}$
can be obtained based on center circle state and vehicle heading
\begin{equation}
s_{k}^{p}=\text{M}^{p}s_{k}^{c}+\textrm{N}^{p},p=\left\{ r,f\right\} ,
\end{equation}
where $\text{M}^{p},\textrm{N}^{p},p=\left\{ r,f\right\} $ denote
linear transformation matrices between consecutive centers. 

Two vehicles $i$ and $j$ are not colliding with each other if none
of their circles overlap. Mathematically, it can be expressed as the
Euclidean distance between two circle centers greater than a threshold
\begin{equation}
\left\Vert \textrm{H}\left(s_{i}^{p_{i}}-s_{j}^{p_{j}}\right)\right\Vert _{2}\geq2\delta,\forall p_{i},p_{j}=\left\{ r,c,f\right\} ,
\end{equation}
\begin{equation}
\text{H=\ensuremath{\left[\begin{array}{cccc}
1 & 0 & 0 & 0\\
0 & 1 & 0 & 0
\end{array}\right]}.}
\end{equation}

Therefore, the SV evasive trajectory planning problem can be formulated
into a quadratic constraint programming as below
\begin{eqnarray}
\min_{\boldsymbol{u}} &  & 1\\
\text{s.t.} 
 &  &\begin{aligned} s_{0}^{c}\left(t+1\right)=\text{A}s_{0}^{c}\left(t\right)+&\text{B}u_{0}\left(t\right),\\ & \forall t=0,...,k-1\label{eq:AV-dynamics},
    \end{aligned}\\
 &  & s_{0}^{p}\left(t\right)=\text{M}^{p}s_{0}^{c}\left(t\right)+\textrm{N}^{p},p=\left\{ r,f\right\} ,\label{eq:Circles-transformation}\\
 &  &\begin{aligned} \left\Vert \textrm{H}\left(s_{0}^{p_{0}}\left(t\right)-s_{i}^{p_{i}}\left(t\right)\right)\right\Vert &_{2}\geq2\delta,\\ & \forall p_{0},p_{i}=\left\{ r,c,f\right\} ,\forall i\label{eq:safety-constraints},
    \end{aligned}\\
 &  & \text{G}u_{0}\left(t\right)\leq\text{h},\label{eq:Kamm-circle}
\end{eqnarray}
where $k$ denotes the number of look-ahead steps, $\boldsymbol{u}=\left[u_{0}\left(0\right),...,u_{0}\left(k-1\right)\right]$ is the SV action sequence and $u_{0}(t)$ denotes the SV action at the $t$-th time step, $s_{0}^{p}(t)$ and $s_{i}^{p}(t),p=\left\{ r,c,f\right\} $
denote the SV and the $i$-th BV state at the $t$-th time step. Note
that the states of all BVs $s_{i}^{p}(t),\forall i,t$ are known since
logged trajectory information is utilized. 
When the remaining time of the observed trajectory is shorter than the look-ahead horizon, we will extend BVs\textquoteright{} trajectories assuming they follow their last moment velocity and heading. 
The decision variables are the action
sequence of the SV in the look-ahead horizon. It is a feasibility
problem since we want to examine whether there exists a safe trajectory
for the SV with respect to all surrounding BVs. Other objectives, such as avoiding large accelerations, or other constraints, such as remaining on the roadway, can be easily added to the objective function or the linear constraints, respectively. The proposed optimization problem can be solved
using existing commercial solvers, for example, Gurobi \cite{gurobi-Gurobi}, which
is used in this study. If the optimization problem is infeasible,
that means the current moment is collision unavoidable for the SV
since there is no feasible trajectory to avoid colliding with other
BVs.

\subsection{Statistical performance analysis using large-scale logged trajectory data \label{subsec:Statistical-performance-analysis}}

As each observed trip is only one realization of the real-world
stochastic driving environment, to unbiasedly and systematically
evaluate the statistical performance of each safety metric, we need to aggregate
the evaluation results from a large number of trips with well-designed statistical measurements.
In this paper, we generate a trajectory dataset using a microscopic traffic simulator, SUMO \cite{lopez2018-SUMO}. The dataset includes both normal driving scenarios and crash
scenarios and the crash type distribution is consistent with that of the
real world. More details of the dataset generation can be found in
Section \ref{subsec:Large-scale-trajectory-dataset}. Two statistical
measurements, i.e., Receiver Operating Characteristic (ROC) curve
and Precision-Recall (PR) curve, are selected and used, which have
different advantages and complement each other.

\begin{figure}
\begin{centering}
\includegraphics[width=0.95\columnwidth]{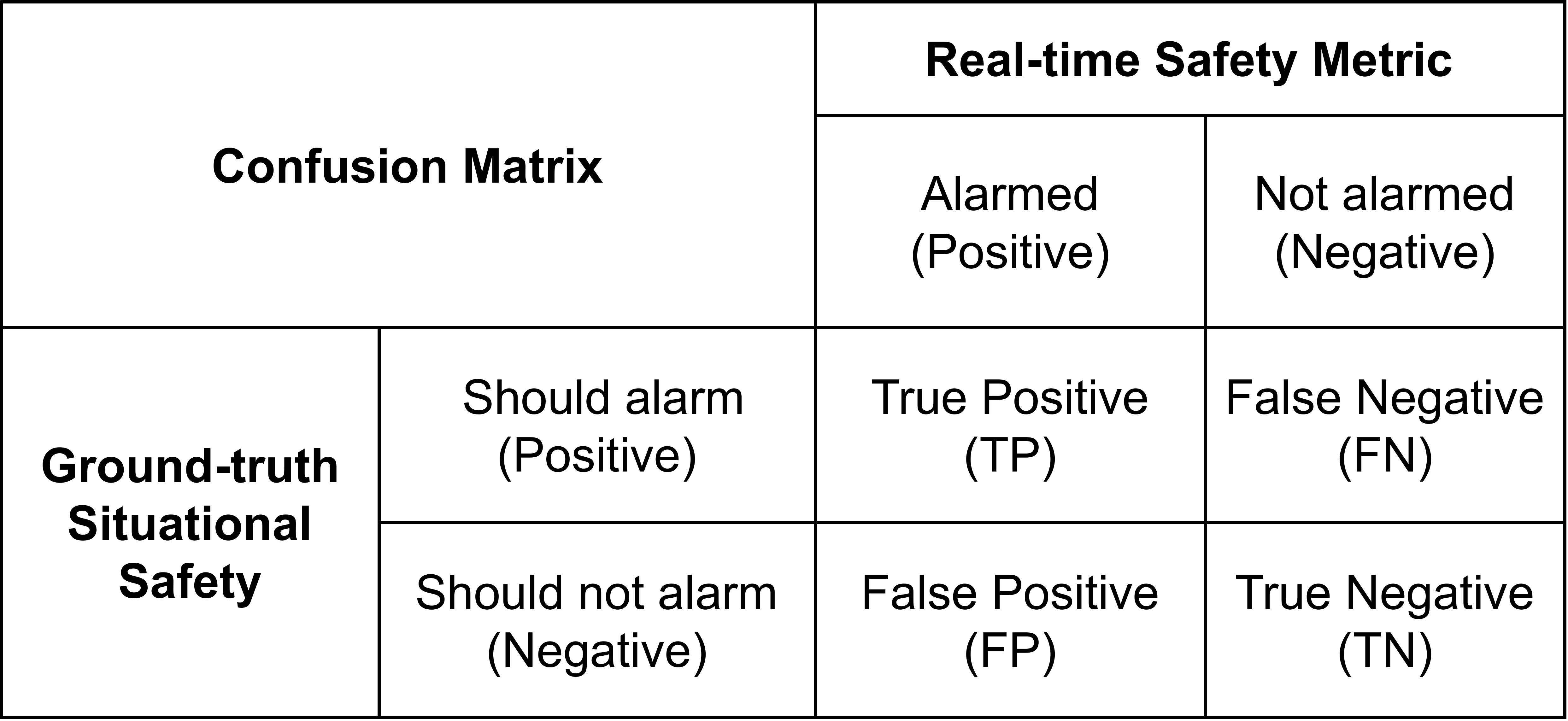}
\par\end{centering}
\caption{Confusion matrix aggregated from large-scale trajectories results.
\label{fig:Confusion-matrix}}
\end{figure}

Based on the metric activation results over logged trajectories, we can construct the confusion
matrix \cite{confusion-matrix-Wiki} as shown in Fig. \ref{fig:Confusion-matrix}. 
There are four
situations, i.e., true positive (TP), false positive (FP), true negative (TN), and
false negative (FN). Two types
of errors exist, i.e., false positive and false negative. False positive
are cases that safety metrics alarm falsely at moments that they should
not. False negative are cases that safety metrics miss to alarm at
dangerous situations. Based on the confusion matrix, the true positive
rate (TPR, which is also called Recall), false positive rate (FPR),
and positive predictive value (PPV, which is also called Precision)
can be calculated \cite{confusion-matrix-Wiki}. The higher the recall and precision, and the lower the FPR, the better
the safety metric performance. These rates will be used to construct
ROC and PR curves.


It has been noted that the safety metric performance is threshold-sensitive
\cite{wang2021review-Review-AAP}. To evaluate the statistical performance
of safety metrics over different activation thresholds, two widely
used measurements, the ROC curve and the PR curve are used. For the
ROC curve, the x-axis is the FPR, and the y-axis is the Recall. The
(0,1) point in the ROC curve represents the best performance a safety
metric can achieve. For the PR curve, the x-axis is the Recall, and
the y-axis is the Precision. The (1,1) point in the PR curve represents
the best performance a safety metric can achieve. Because each point
in the ROC and PR curves are calculated based on a specific activation
threshold, by using ROC and PR curves, we can study the performance
of a safety metric with different values of its activation threshold,
which can systematically reflect the performance of a safety metric.

The ROC curve is not sensitive to changes in class distribution in
the test dataset \cite{fawcett2006introduction-ROC}. That means
if the proportion of positive (should alarm) and negative (should
not alarm) instances changes in the evaluated large-scale trajectory
dataset, the ROC curve remains the same. As a result, the ROC can
produce a robust measurement even if the test dataset has distributional
biases with the real-world situation. To compare the performance over
multiple safety metrics, a widely used method is to calculate the
area under the curve (AUC).
AUC provides an aggregate measure of different metrics: the larger
the AUC, the better performance.

Comparing with the ROC curve, when dealing with highly skewed datasets,
the PR curve can give a more informative picture of a metric's performance
\cite{davis2006relationship-ROC-vs-PR}. That is exactly the case
in our application. In the real-world driving environment, the positive
cases, i.e., moments that are dangerous and a safety metric should
alarm, are significantly less than negative cases. If a safety metric
alarms aggressively, it might achieve good performance in the ROC
curve as it can obtain high recall and also low FPR since the number
of negative moments is dominant. As a result, it might false-alarm
frequently in safe situations. To alleviate this effect, the PR curve
considers the precision to quantify the percentage of correctly alarmed
cases among all alarmed cases. Note that there usually exists
a trade-off between precision and recall. Because the improvement
of the recall rate often time leads to an increase of false alarm
cases, which decreases the precision. Therefore, the PR curve can
also be used to identify the optimal activation threshold, balancing
the recall and the precision. 

\section{Case studies \label{sec:Sec-3}}

In this section, we demonstrate the capability of the proposed framework
for the performance evaluation of real-time safety metrics. Three
representative real-time safety metrics are selected, including TTC
\cite{hayward1972near-TTC}, PCM \cite{junietz2018criticality-PCM},
and MPrISM \cite{weng2020-MPrISM}. A large-scale simulated trajectory
dataset is first generated which includes both normal and collision
trajectories. Then, we conduct failure analysis of real-time safety
metrics. Four example scenarios that real-time safety metrics failed
are demonstrated and analyzed. The statistical performance
analysis is followed to comprehensively evaluate and compare different
metrics.
Finally, we use a real-world event to demonstrate the proposed method is applicable to the real-world dataset.

\subsection{Three evaluated real-time safety metrics \label{subsec:Introduction-of-3-metrics}}

In this section, we will briefly introduce the three selected metrics.
The first metric is the TTC \cite{hayward1972near-TTC} which has
been widely used in both industry and academia. It is defined by
\begin{equation}
TTC=\frac{x_{i-1}-x_{i}-d}{v_{i}-v_{i-1}},
\end{equation}
where $x_{i-1}$ and $v_{i-1}$ denote the leading vehicle longitudinal
position and speed, respectively. $x_{i}$ and $v_{i}$ denote corresponding
values for the following vehicle. $d$ denotes the vehicle length. 

The second metric is the PCM \cite{junietz2018criticality-PCM} proposed
from the PEGASUS project \cite{PEGASUS-PEGASUS}. The situational safety evaluation task is
modeled as a SV trajectory planning problem using model predictive control (MPC). BVs are assumed
to maintain their current velocity and heading during the prediction
horizon and the decision variables are the action sequences of the
SV. The objective is to minimize the criticality within the prediction
horizon. Specifically, the optimization problem is formulated as 

\begin{eqnarray}
\min_{\boldsymbol{u}} 
 &  & \begin{aligned} \sum_{t=1}^{k-1}(w_{x}R_{x}\left(t\right)+w_{y}&R_{y}^{2}\left(t\right)+w_{ax}\frac{a_{x}^{2}\left(t\right)}{\left(\mu_{max}g\right)^{2}} \\ &+w_{ay}\frac{a_{y}^{2}\left(t\right)}{\left(\mu_{max}g\right)^{2}})
 \end{aligned}\\
\text{s.t.}
 &  & \begin{aligned} x\left(t+1\right)=A\left(t\right)x\left(t\right)&+B\left(t\right)u\left(t\right),\\ &\forall t=0,...,k-1\label{eq:PCM-system-dynamics},
    \end{aligned}\\
 &  & c_{r}\left(t\right)\leq y\left(t\right)\leq c_{l}\left(t\right),\label{eq:PCM-y-bound}\\
 &  & x\left(t\right)\leq c_{f}\left(t\right),\label{eq:PCM-x-bound}\\
 &  & Gu\left(t\right)\leq h,\label{eq:PCM-Kamm-1}
\end{eqnarray}
The criticality is composed of four components, i.e., the longitudinal
margin $R_{x}\left(t\right)$, the lateral margin $R_{y}\left(t\right)$,
the longitudinal acceleration $a_{x}\left(t\right)$, and the lateral
acceleration $a_{y}\left(t\right)$. $w_{x},w_{y},w_{ax},$ and $w_{ay}$
denote pre-defined weighting factors of each component. $u(t)$ denotes
the SV action at the $t$-th timestep which includes $a_{x}\left(t\right)$
and $a_{y}\left(t\right)$. $k$ denotes the look-ahead steps and
$\mu_{max}$ denotes the maximum available friction coefficient. Eq.
(\ref{eq:PCM-system-dynamics}) denotes the system dynamics of the SV
and $s\left(k\right)$ denotes the SV state at the $t$-th timestep.
Eqs. (\ref{eq:PCM-y-bound}-\ref{eq:PCM-x-bound}) denote the safety
constraints where $c_{l},c_{r},$ and $c_{f}$ represent the left,
right, and front boundaries, respectively, which depend on the predicted
trajectories of surrounding vehicles. Eq. (\ref{eq:PCM-Kamm-1}) denotes
the admissible action space according to the Kamm's circle. By solving
the SV trajectory optimization problem at each time step, we can obtain
the safety metric using either the criticality value or the maximum
estimated acceleration during the prediction horizon.

The third metric is the MPrISM \cite{weng2020-MPrISM} proposed by National Highway Traffic Safety Administration (NHTSA). The situational safety
is evaluated by considering the pairwise interaction between the SV
and each BV. It assumes there exists one non-cooperative BV, denoted
as the principal other vehicle (POV), which will take the worst action
to create the safety-critical situation for the SV, while all other
BVs will comply with the SV to avoid collision. Therefore, at a snapshot
with the state of SV, $x_{0}$, and the state of POV, $x_{i}$, for
each look-ahead step $N=1,...,T$, the relative distance between the
SV and the POV at the end of the prediction horizon $N$ can be obtained
by solving the following minimax problem:
\begin{eqnarray}
 \min_{\boldsymbol{u_{i}}}\max_{\boldsymbol{u_{0}}} &  & \left\Vert x_{i}\left(N\right)-x_{0}\left(N\right)\right\Vert \\
\text{s.t.} 
 &  & \begin{aligned} x_{0}\left(t+1\right)=A_{0}\left(t\right)&x_{0}\left(t\right)+B_{0}\left(t\right)u_{0}\left(t\right),\hspace{15pt}\\ &\forall t=0,...,N-1\label{eq:MPrISM-system-dynamics-SV}
     \end{aligned}\\
 &  & \begin{aligned} x_{i}\left(t+1\right)=A_{i}\left(t\right)&x_{i}\left(t\right)+B_{i}\left(t\right)u_{i}\left(t\right),\hspace{15pt}\\ &\forall t=0,...,N-1 &\label{eq:MPrISM-system-dynamics-POV}
     \end{aligned}\\
 &  & G_{0}u_{0}\left(t\right)\leq h_{0},\label{eq:MPrISM-Kamm-1-SV}\\
 &  & G_{i}u_{i}\left(t\right)\leq h_{i},\label{eq:MPrISM-Kamm-1-POV}
\end{eqnarray}
where $\boldsymbol{u_{0}}$ and $\boldsymbol{u_{i}}$ denote actions
sequences of the SV and the POV, respectively. Eq. (\ref{eq:MPrISM-system-dynamics-SV})
and Eq. (\ref{eq:MPrISM-Kamm-1-SV}) denotes the system dynamics and
the Kamm circle constraints of the SV, respectively. Eq. (\ref{eq:MPrISM-system-dynamics-POV})
and Eq. (\ref{eq:MPrISM-Kamm-1-POV}) denotes the corresponding constraints
for the POV. If the distance at the end of the prediction horizon
$N\Delta$ ($\Delta$ is the time resolution) is smaller than a pre-determined
collision threshold $C$, the time-to-collision with respect to the
specific POV equals to $N\Delta$. Therefore, after looping over all
BVs, the least time-to-collision value can be obtained and denoted
as the model predictive time-to-collision (MPrTTC), which is used
as the safety metric for the snapshot.

The model settings and parameters of both PCM and MPrISM are set according
to their original papers. The detailed settings and parameters of
the three safety metrics and the ground-truth situational safety calculation can be found
in Appendix. In failure analysis, we choose metrics activation
thresholds as follows. TTC is considered alarmed if it is smaller
than 1 second, MPrISM is considered alarmed if the MPrTTC is smaller
than 1 second, and the PCM is considered alarmed if the maximum expected
acceleration equals to or greater than 8 $m/s^{2}$. In statistical
performance analysis, a set of varying activation thresholds will
be used for each metric. For the MPrISM and TTC, the varying thresholds
range from $0.1s$ to $4.0s$ with a 0.1 step size. For the PCM, the
varying thresholds range from $0.8m/s^{2}$ to $8.0m/s^{2}$ with
a 0.8 step size, which are corresponding to 10\% to 100\% vehicle
maximum deceleration.

\subsection{Trajectory dataset \label{subsec:Large-scale-trajectory-dataset}}

In this study, we generate the trajectory dataset using a straight three-lane highway environment
modeled in SUMO. The length and width of all vehicles
are 5m and 2m, respectively. The lane width is 4m. The SV model is
developed using the double deep Q-networks (DDQN) algorithm \cite{van2016deep-DDQN}
with dueling networks \cite{wang2016-Dueling}, considering
both safety and mobility \cite{feng2021intelligent-NADE}. 

As discussed, since each observed trip is a realization of the real-world
stochastic driving environment, we need to use a large-scale trajectory
dataset to systematically analyze the safety metric performance. The dataset needs to include vehicle state information for both SV and surrounding BVs at each timestep since they are required
as safety metrics inputs. It should also cover
both normal and safety-critical driving situations according to the probabilistic distribution in the real world. To the best of our knowledge, currently 
no suitable dataset is available to use. Therefore, we generate a simulated
trajectory dataset. To reproduce a naturalistic driving
environment (NDE), the BV driving behaviors are modeled using real-world
naturalistic driving data (NDD) from the Safety Pilot Model Deployment
project \cite{bezzina2014-SPMD}. More details of the NDE modeling
method used in this study can be found in \cite{yan2021distributionally-NDE}.
In order to increase the efficiency of generating safety-critical
trajectories, the naturalistic and adversarial driving environment
(NADE) \cite{feng2021intelligent-NADE} was used to generate the trajectories with collision. 

\begin{figure}
\begin{centering}
\includegraphics[width=0.8\columnwidth]{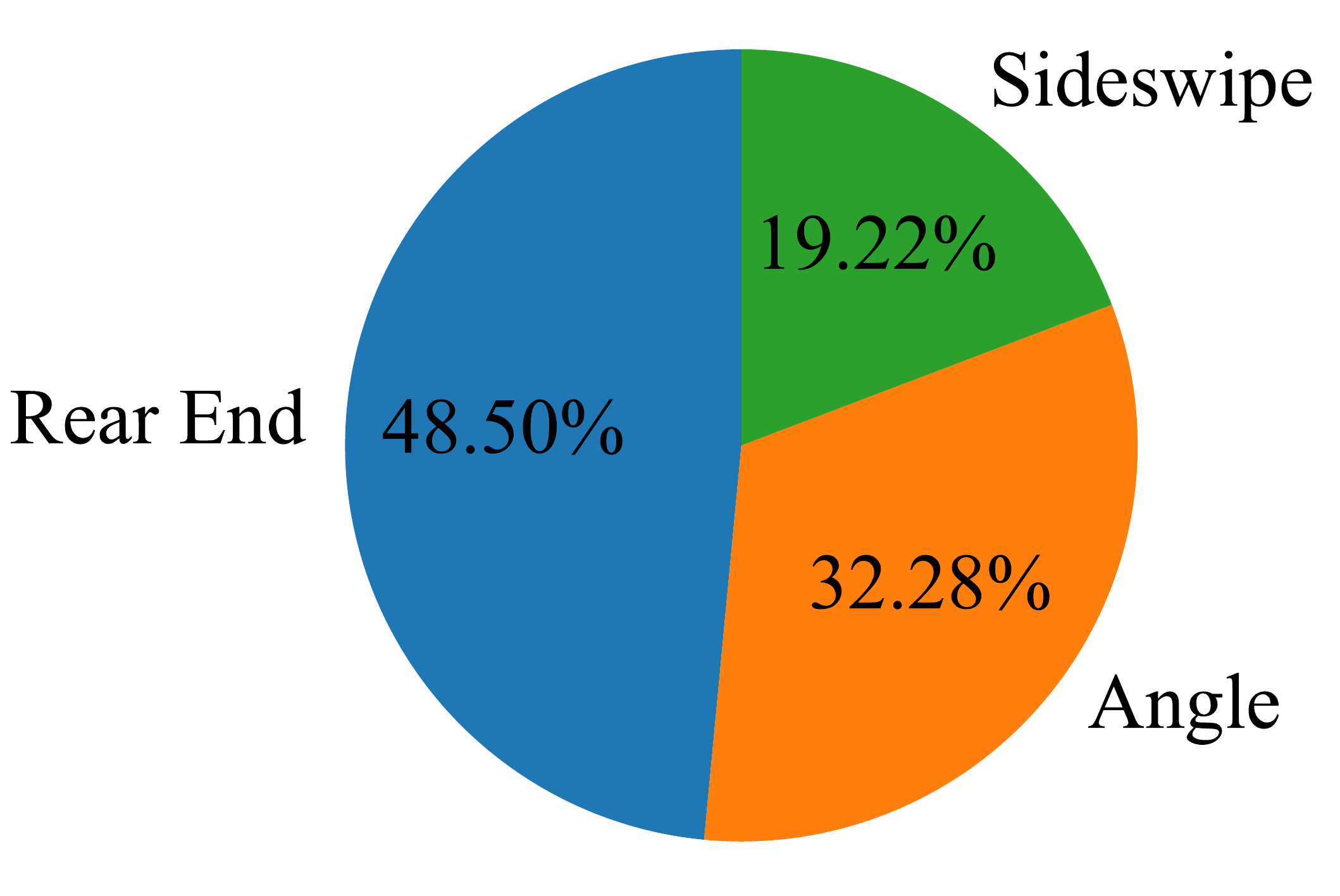}
\par\end{centering}
\caption{Crash type proportion in the simulated dataset. \label{fig:Crash-type-proportion}}
\end{figure}

The trajectory dataset includes 5,050 simulated trajectories, 5000
of which have no crashes and the rest have crashes. In each trajectory,
the SV drives $400m$ on the highway. Therefore, the simulated dataset
covers over $2,000km$ ego-vehicle travel distances. Different crash
types based on the first harmful event including rear-end, angle,
and sideswipe are involved in the dataset. The
proportions of different crash types in the dataset are shown in Fig.
\ref{fig:Crash-type-proportion}. It is consistent with the real-world crash type distribution reported by NHTSA \cite{NHTSA-safety-facts}.

\subsection{Failure analysis of real-time safety metrics}

In this subsection, four example scenarios that the real-time safety metrics
may fail are demonstrated and analyzed: SV overtakes BV, BV cuts in
SV, BV and SV both move into the same lane, and BV lane-changes to
SV adjacent lane. Using the proposed framework, the failure analysis
can effectively analyze the effects of metric assumptions, approximations,
and parameters, which provides valuable insights for potential refinements
and improvements of metrics. 

\subsubsection{Scenario 1: SV overtakes BV}

\begin{figure}
\begin{centering}
\subfloat[\label{fig:Scenario1-origin-traj}]{\begin{centering}
\includegraphics[width=0.9\columnwidth]{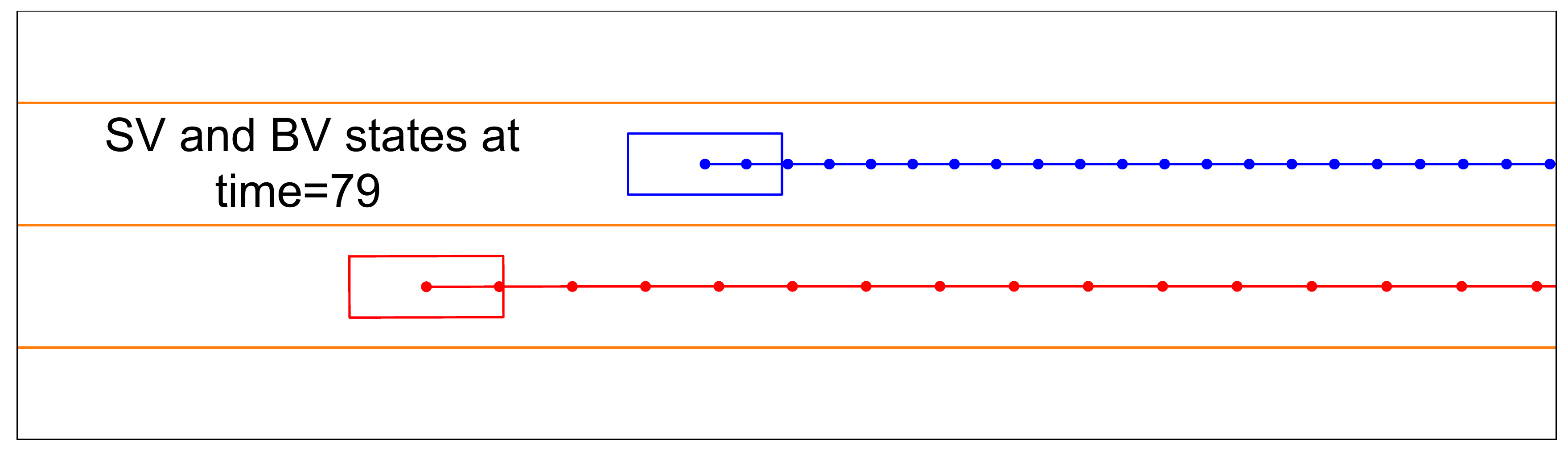}
\par\end{centering}
}
\par\end{centering}
\centering{}\subfloat[\label{fig:Scenario1-MPrISM-traj}]{\begin{centering}
\includegraphics[width=0.9\columnwidth]{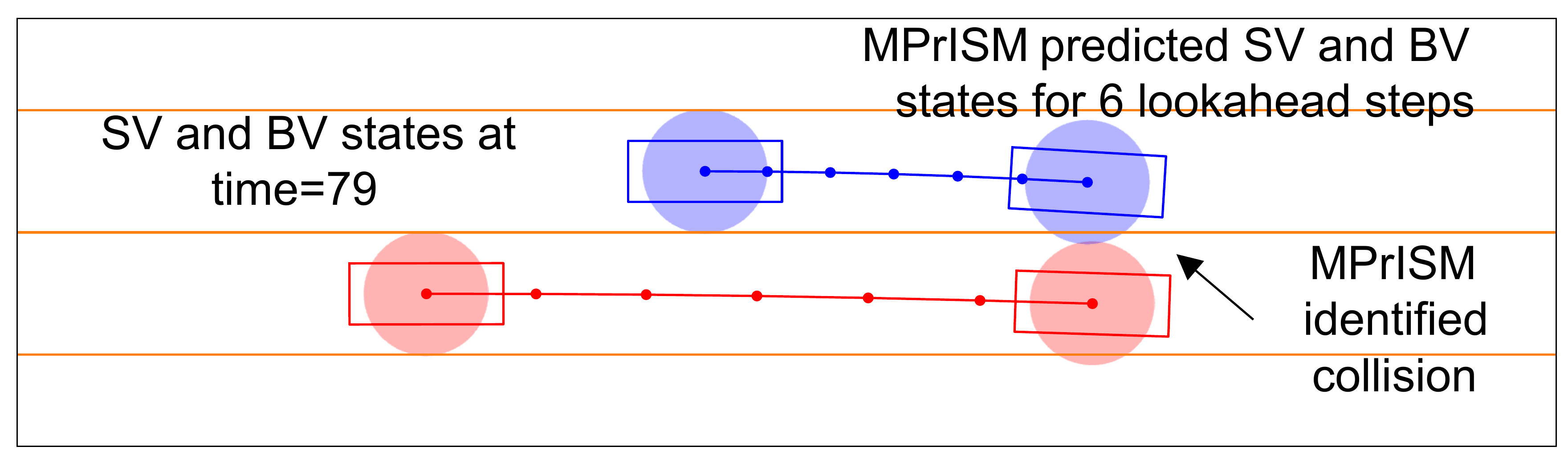}
\par\end{centering}
}\caption{(a) the logged SV (Red) and POV (Blue) trajectories starting from
timestep 79, (b) the MPrISM predicted SV (Red) and POV (Blue) trajectories
starting from timestep 79 of Scenario 1. The shaded area
denotes the assumed vehicle geometry by the safety metric.\label{fig:Scenario1-traj}}
\end{figure}

\begin{figure}
\begin{centering}
\includegraphics[width=1.0\columnwidth]{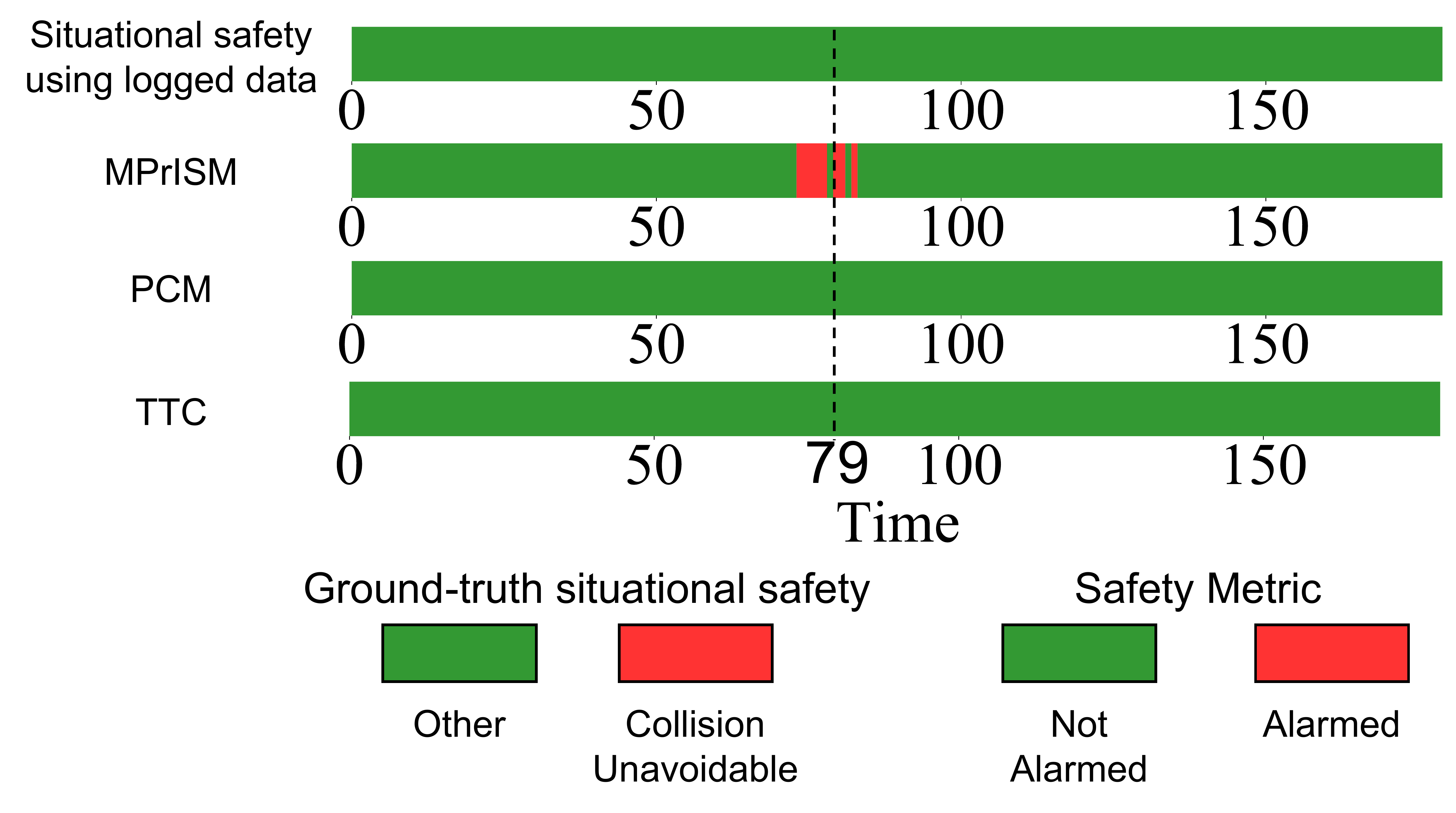}
\par\end{centering}
\caption{The situational safety calculated using logged trajectory data and safety metrics results of Scenario 1.
\label{fig:Scenario1-res}}
\end{figure}

In the first scenario, the SV overtakes a BV on the adjacent lane
and there is no crash happens. The logged trajectories of SV (shown in red) and
BV (shown in blue) starting from timestep 79 are shown in Fig. \ref{fig:Scenario1-origin-traj}.
The safety metrics results and the situational safety calculated using logged trajectory data are shown in Fig. \ref{fig:Scenario1-res}. The red color moments
in the ground-truth situational safety calculation denote the moments that are collision
unavoidable. The red color moments in the safety metric denote the
moments that the metric alarms. The MPrISM produces false-positive
cases in this scenario. 

From logged vehicle trajectory calculation results, we can find that the moment 79
is not dangerous, and all metrics except for the MPrISM do not alarm.
The MPrISM predicted trajectories of SV and BV are shown in Fig.
\ref{fig:Scenario1-MPrISM-traj}. The shaded area denotes the assumed
vehicle geometry by MPrISM. The BV is trying to steer into the SV\textquoteright s
lane since MPrISM assumes the BV will take the worst-case behaviors.
The MPrTTC is 0.6s since a collision is predicted to happen after
6 look-ahead steps. According to the collision definition in MPrISM,
a collision is considered to happen when the Euclidean distance between
two vehicles (considered as mass points with a certain radius) is
smaller than the collision threshold. From the results as shown in
Fig. \ref{fig:Scenario1-MPrISM-traj}, we can find the predicted
SV and BV circles overlap, so an accident is predicted by MPrISM.
However, the two vehicles are not colliding with each other if considering
their real geometry as shown by the two rectangles. Therefore, a false-positive
case is produced by the MPrISM due to both the worst-case behavior
assumption and the single circle approximation. There is a trade-off
for setting the MPrISM collision threshold. Using a larger threshold
will cause more false alarm cases as shown in this scenario, whereas
using a smaller threshold will cause false-negative cases (i.e., miss
dangerous situations) as illustrated in Scenario 3.

\subsubsection{Scenario 2: BV cuts in SV}

\begin{figure}
\begin{centering}
\subfloat[\label{fig:Scenario2-origin-traj}]{\begin{centering}
\includegraphics[width=0.9\columnwidth]{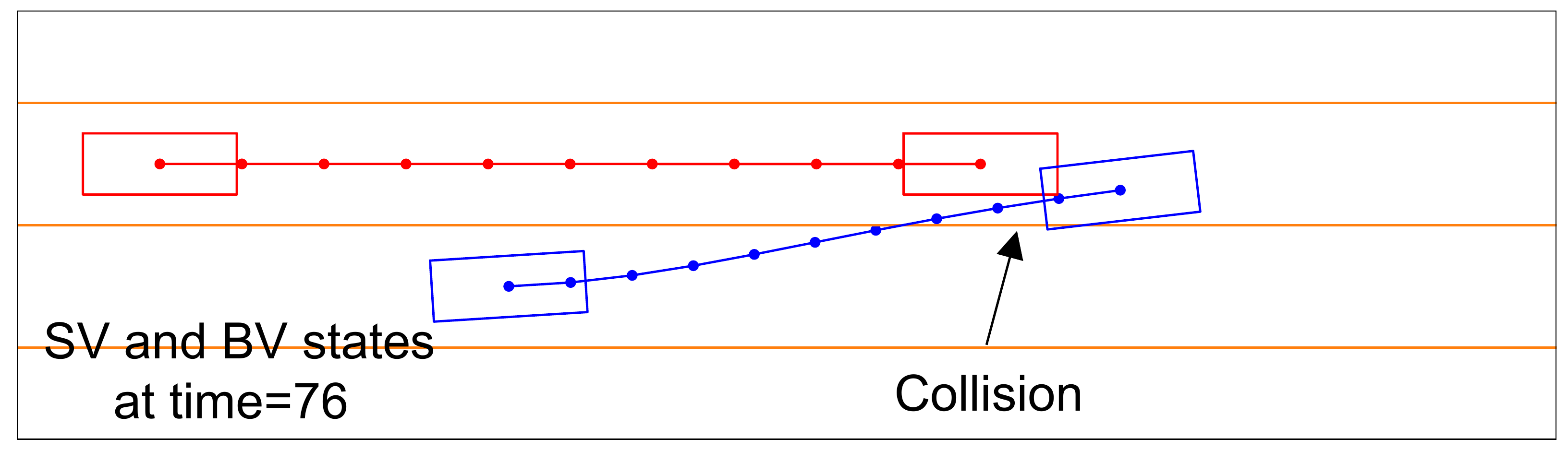}
\par\end{centering}
}
\par\end{centering}
\centering{}\subfloat[\label{fig:Scenario2-PCM-traj}]{\begin{centering}
\includegraphics[width=0.9\columnwidth]{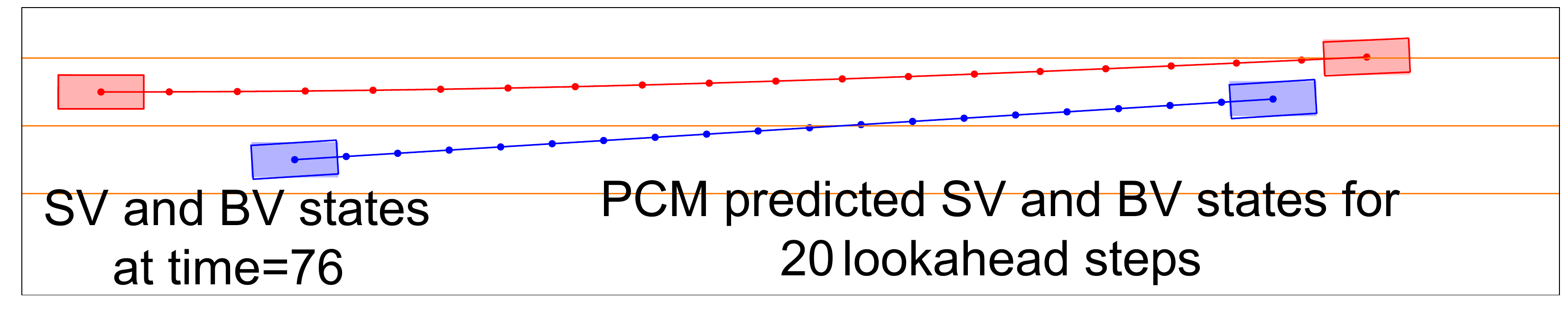}
\par\end{centering}
}\caption{(a) the logged SV (Red) and POV (Blue) trajectories starting from
timestep 76, (b) the PCM predicted SV (Red) and POV (Blue) trajectories
starting from timestep 76 of Scenario 2. The shaded area
denotes the assumed vehicle geometry by the safety metric. \label{fig:Scenario2-traj}}
\end{figure}

\begin{figure}
\begin{centering}
\includegraphics[width=1.0\columnwidth]{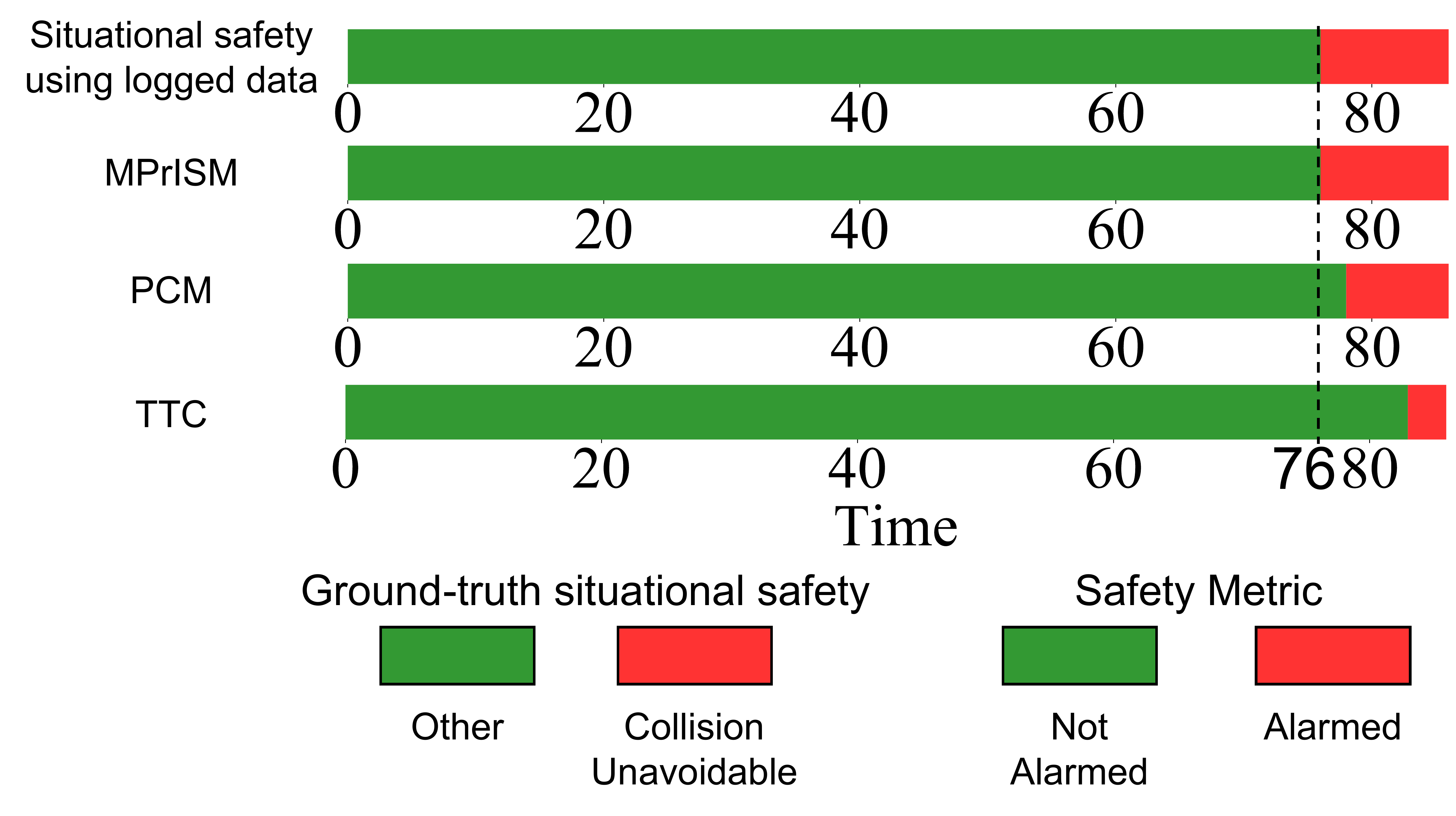}
\par\end{centering}
\caption{The situational safety calculated using logged trajectory data and safety metrics results of Scenario 2.
\label{fig:Scenario2-res}}
\end{figure}

In the second scenario, the BV cuts in the SV and a crash happened
during the process. The logged trajectories of SV (shown in red) and BV (shown
in blue) starting from timestep 76 are shown in Fig. \ref{fig:Scenario2-origin-traj}.
The safety metrics results and the situational safety calculated using logged trajectory data are shown in
Fig. \ref{fig:Scenario2-res}. The PCM and TTC produce false-negative cases in
this scenario. 

From logged vehicle trajectory calculation results, we can find that starting from
moment 76, the SV is collision unavoidable. The MPrISM successfully
alarms starting from the moment 76. However, both PCM and TTC fail
to activate at the moment and cause a false negative. For the PCM,
its predicted SV and BV trajectories are shown in Fig. \ref{fig:Scenario2-PCM-traj}.
The PCM assumes BV will maintain its current speed and heading within
the prediction horizon. As a result, the SV can avoid the crash by
turning to the adjacent lane. However, in the real situation, the
BV is cutting in more aggressively compared with PCM behavior assumptions.
Therefore, the PCM fails to activate at this collision unavoidable
moment. 

For the TTC, it fails to consider the cutting in BV since they are
not in the same lane at the current moment. Thus, the SV does not
consider the BV as its leading vehicle and produces false-negative
cases. 

\subsubsection{Scenario 3: BV and SV both move into the same lane}

\begin{figure}
\begin{centering}
\subfloat[\label{fig:Scenario3-origin-traj}]{\begin{centering}
\includegraphics[width=0.9\columnwidth]{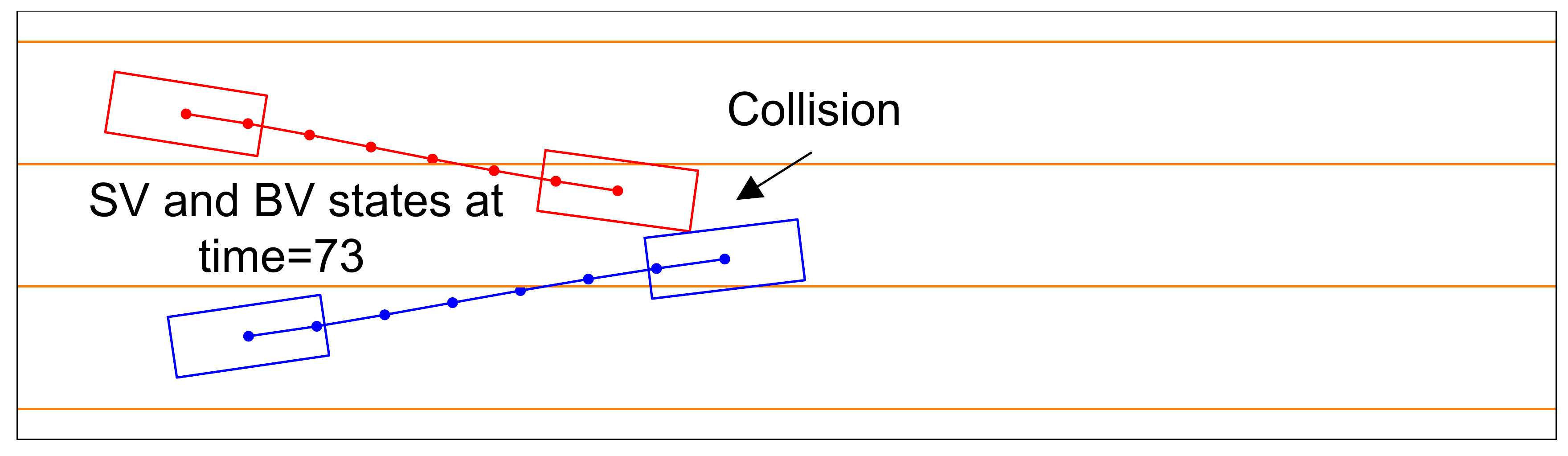}
\par\end{centering}
}
\par\end{centering}
\centering{}\subfloat[\label{fig:Scenario3-MPrISM-traj}]{\begin{centering}
\includegraphics[width=0.9\columnwidth]{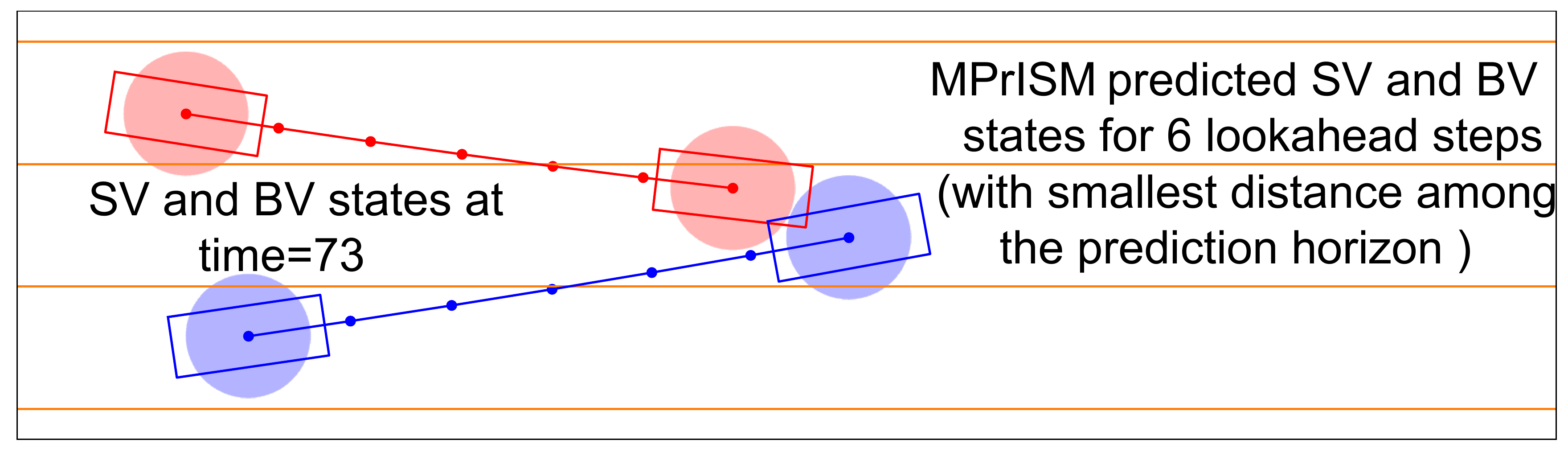}
\par\end{centering}
}\caption{(a) the logged SV (Red) and POV (Blue) trajectories starting from
timestep 73, (b) the MPrISM predicted SV (Red) and POV (Blue) trajectories
starting from timestep 73 of Scenario 3. The shaded area
denotes the assumed vehicle geometry by the safety metric.\label{fig:Scenario3-traj}}
\end{figure}

\begin{figure}
\begin{centering}
\includegraphics[width=1.0\columnwidth]{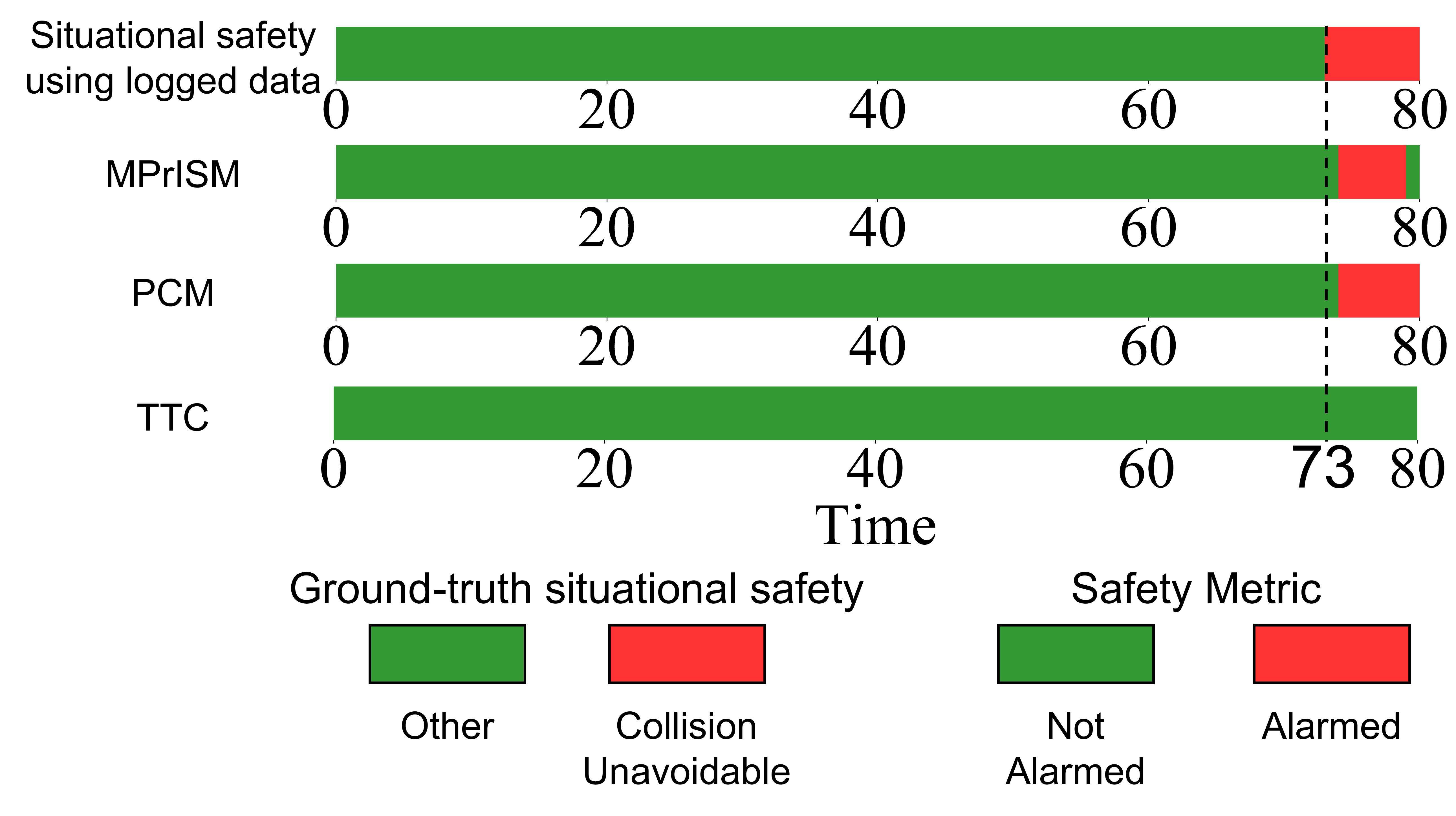}
\par\end{centering}
\caption{The situational safety calculated using logged trajectory data and safety metrics results of Scenario 3.
\label{fig:Scenario3-res}}
\end{figure}

In the third scenario, the BV and SV are lane-changing to the same
target lane and a crash happens during the process. The logged trajectories of
SV (shown in red) and BV (shown in blue) starting from timestep 73
are shown in Fig. \ref{fig:Scenario3-origin-traj}. The safety metrics results and the situational safety calculated using logged trajectory data are shown in
Fig. \ref{fig:Scenario3-res}. All three metrics
produce false-negative cases in this scenario. 

From logged vehicle trajectory calculation results, we can find that starting from
moment 73, the SV is collision unavoidable. However, all three metrics
fail to alarm before the moment and cause a false
negative. The MPrISM predicted trajectories of SV and POV are shown
in Fig. \ref{fig:Scenario3-MPrISM-traj}. Within the 10 looking-ahead
steps of MPrISM, the minimum predicted distance between the SV and
BV occurs at the number 6 look-ahead steps. As shown in Fig. \ref{fig:Scenario3-MPrISM-traj},
the two vehicles have already collided with each other as shown by
the rectangles. However, due to the single circle approximation adopted
by the MPrISM, it fails to identify the collision. The benefit of
using three circles rather than one circle is that it can more accurately
capture the vehicle geometry and determine whether a collision happens
or not. For the PCM and TTC, the reasons why they fail in this scenario
are similar as discussed in Scenario 2. Another reason why TTC fails
in this scenario is that the leading vehicle velocity is higher than
the following vehicle velocity.

\subsubsection{Scenario 4: BV lane-change to SV adjacent lane}

\begin{figure}
\begin{centering}
\subfloat[\label{fig:Scenario4-origin-traj}]{\begin{centering}
\includegraphics[width=0.9\columnwidth]{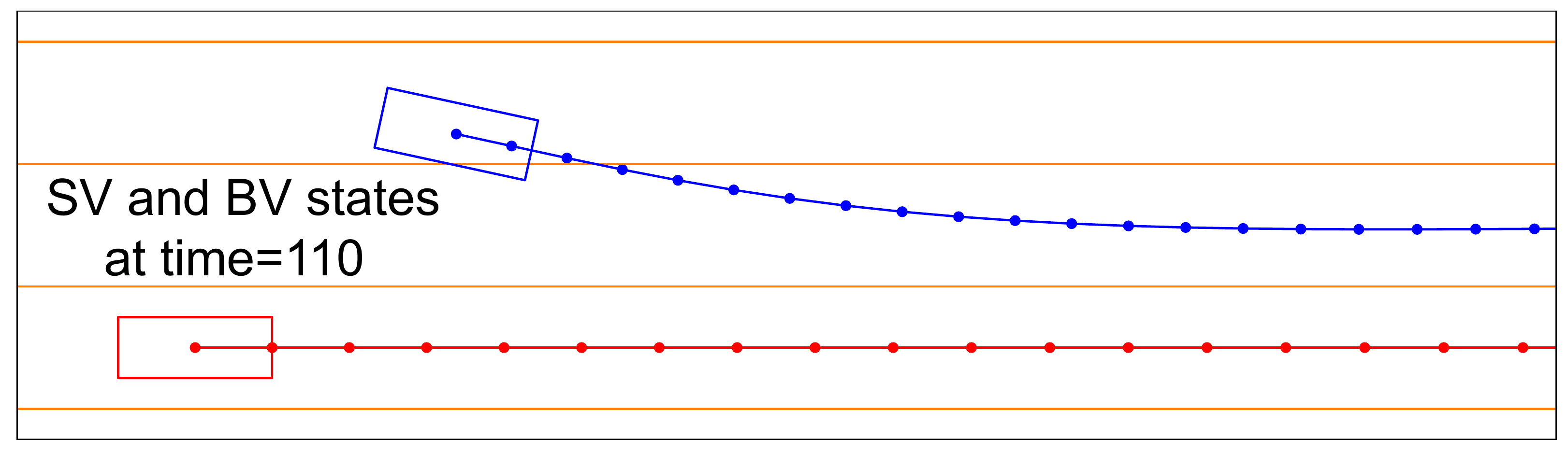}
\par\end{centering}
}
\par\end{centering}
\centering{}\subfloat[\label{fig:Scenario4-MPrISM-traj}]{\begin{centering}
\includegraphics[width=0.9\columnwidth]{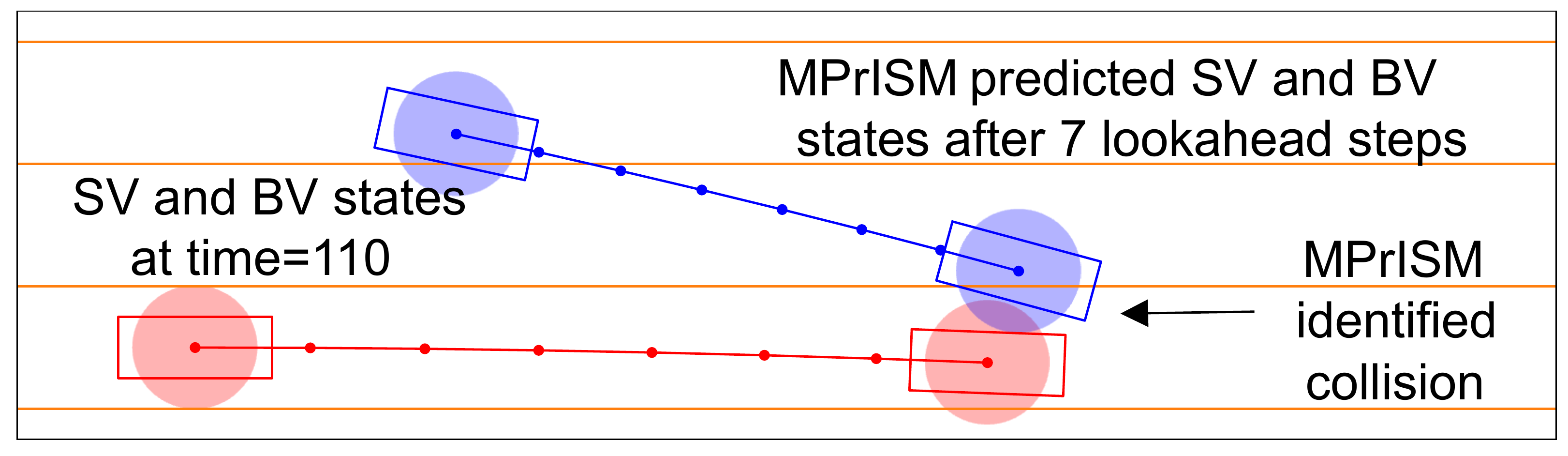}
\par\end{centering}
}\caption{(a) the logged SV (Red) and POV (Blue) trajectories starting from
timestep 110, (b) the MPrISM predicted SV (Red) and POV (Blue) trajectories
starting from timestep 110 of Scenario 4. The shaded area
denotes the assumed vehicle geometry by the safety metric.\label{fig:Scenario4-traj}}
\end{figure}

\begin{figure}
\begin{centering}
\includegraphics[width=1.0\columnwidth]{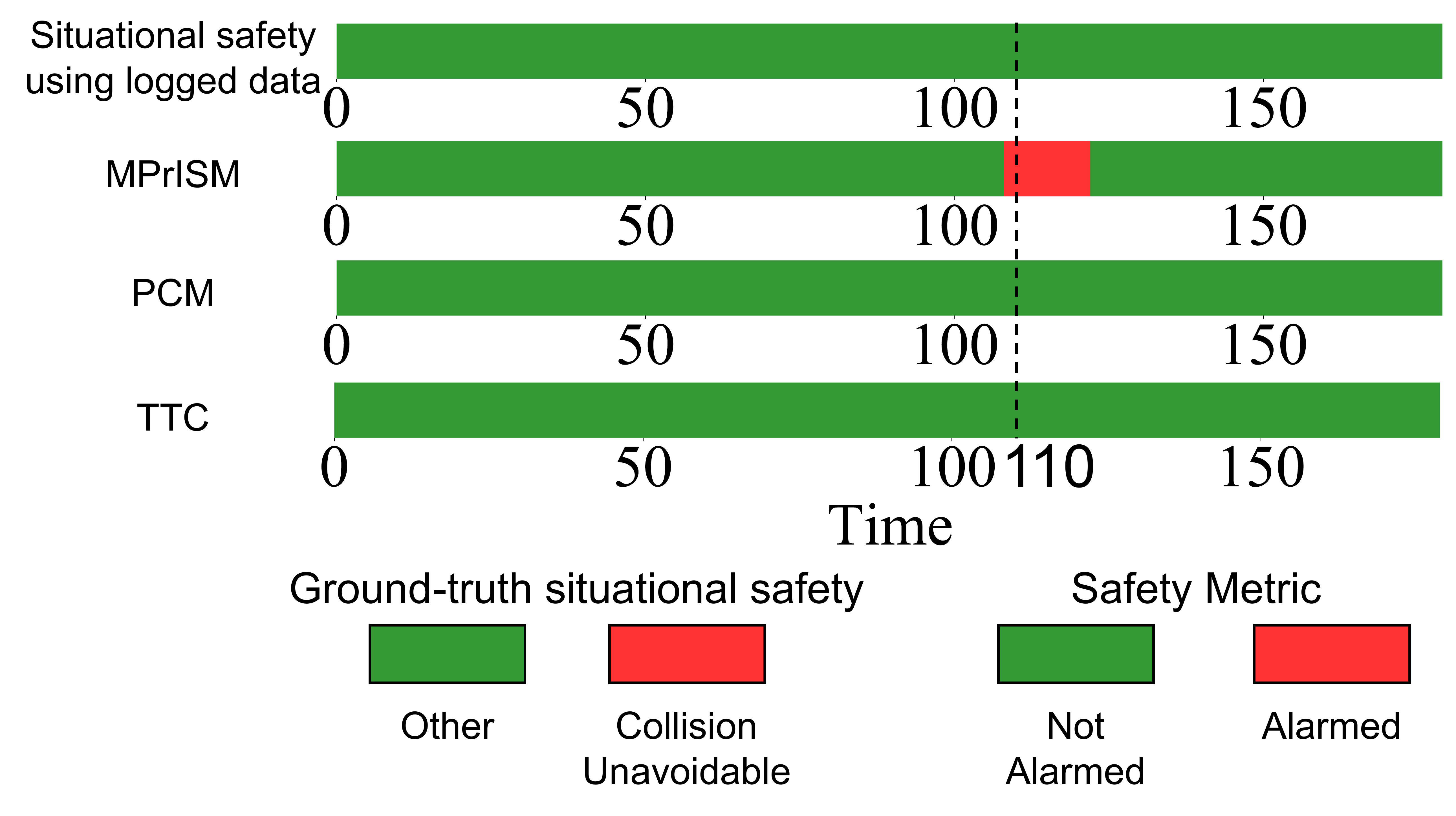}
\par\end{centering}
\caption{The situational safety calculated using logged trajectory data and safety metrics results of Scenario 4.
\label{fig:Scenario4-res}}
\end{figure}

In the fourth scenario, the BV makes a lane change to the SV adjacent
lane and there is no crash. The logged trajectories of SV (shown in red) and BV (shown
in blue) starting from timestep 110 are shown in Fig. \ref{fig:Scenario4-origin-traj}.
The safety metrics results and the situational safety calculated using logged trajectory data are shown in
Fig. \ref{fig:Scenario4-res}. The MPrISM produces false-positive cases in
this scenario. 

\begin{figure*}
        \centering
 
        \subfloat[\label{fig:Safety-metrics-performance-varying-0}]{
            \begin{minipage}[b]{0.32\textwidth}
            \includegraphics[width=1\textwidth]{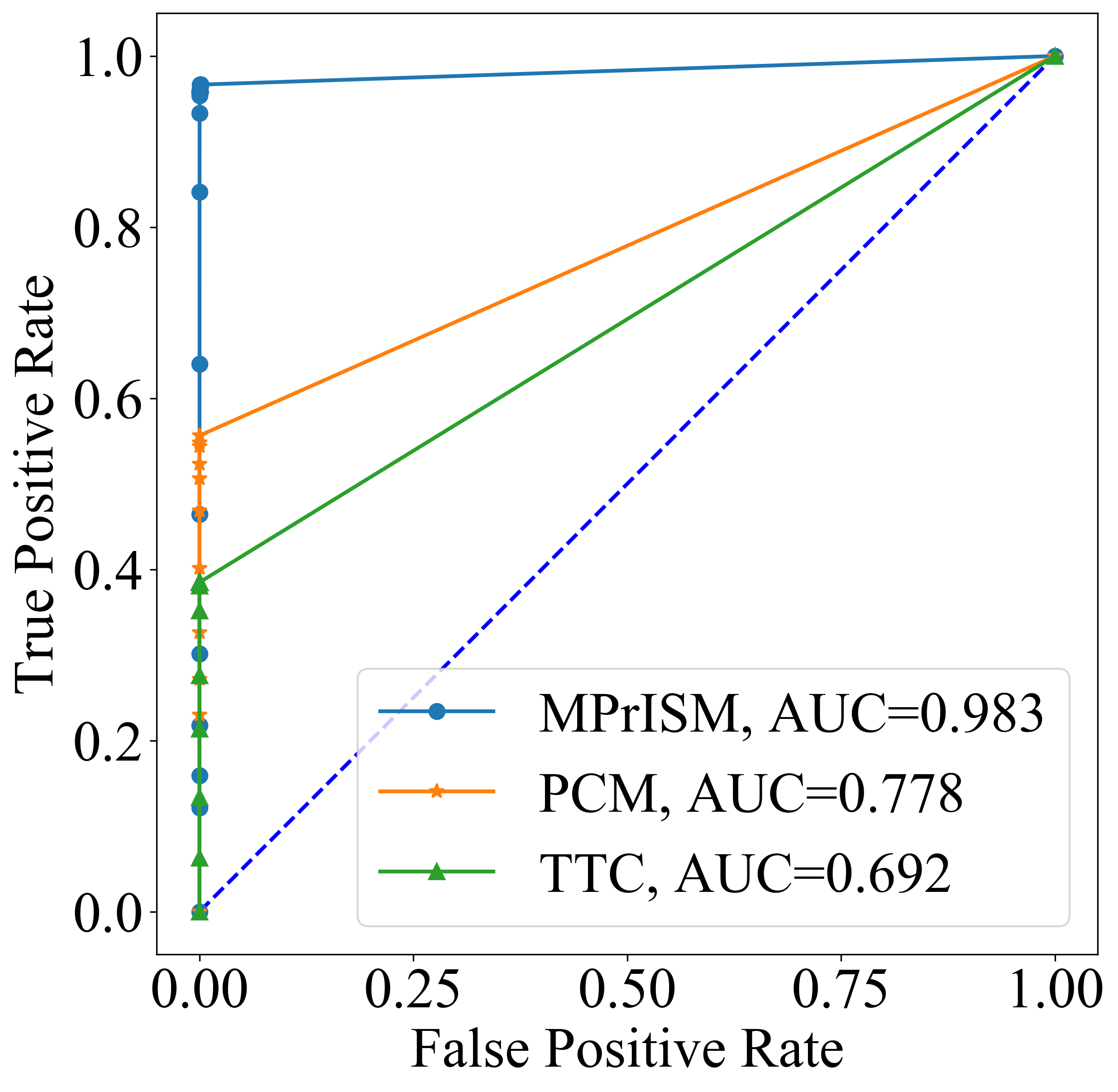} \\
            \includegraphics[width=1\textwidth]{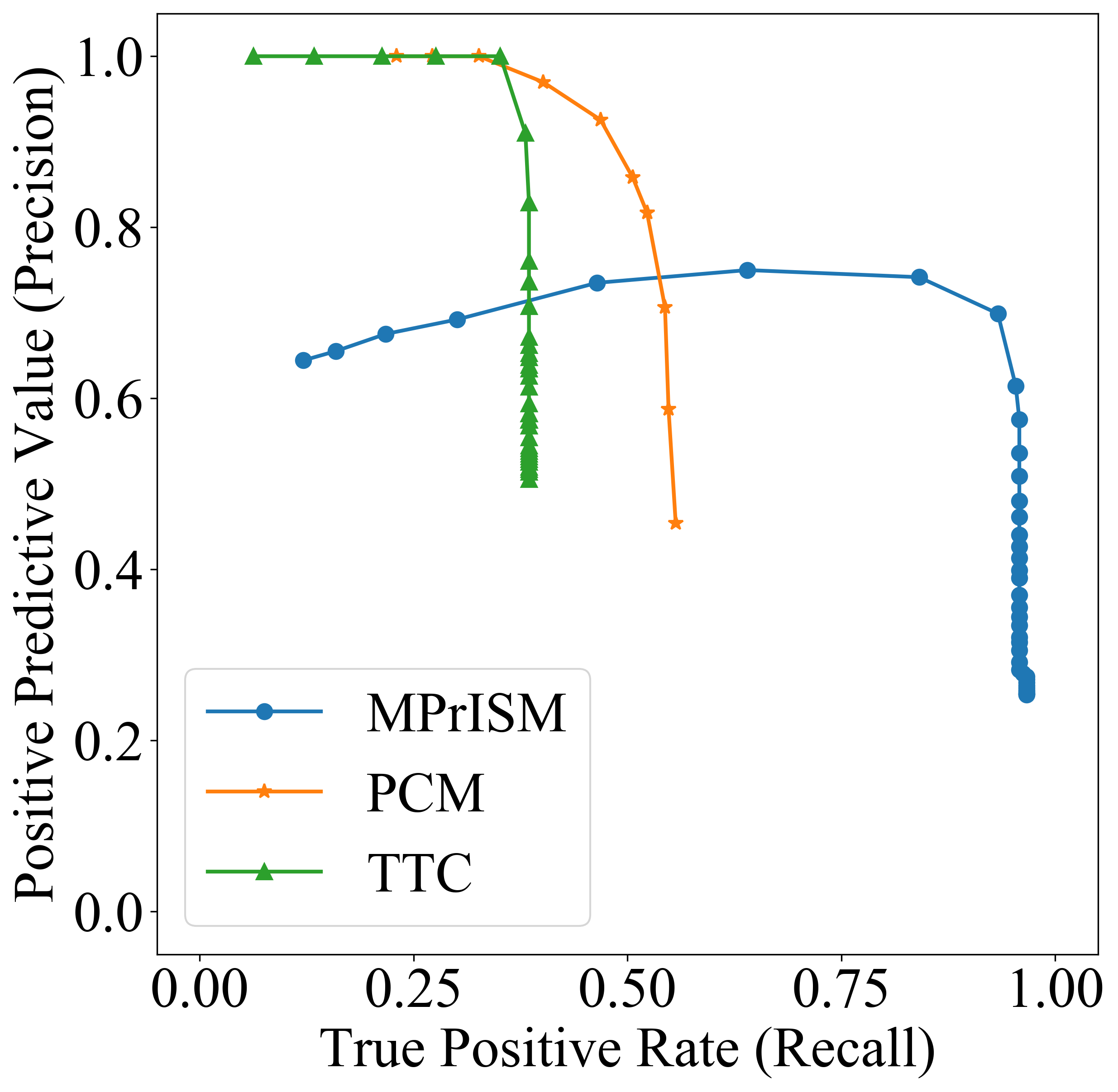}
            \end{minipage}
            }
        \subfloat[\label{fig:Safety-metrics-performance-varying-0.5}]{
            \begin{minipage}[b]{0.32\textwidth}
            \includegraphics[width=1\textwidth]{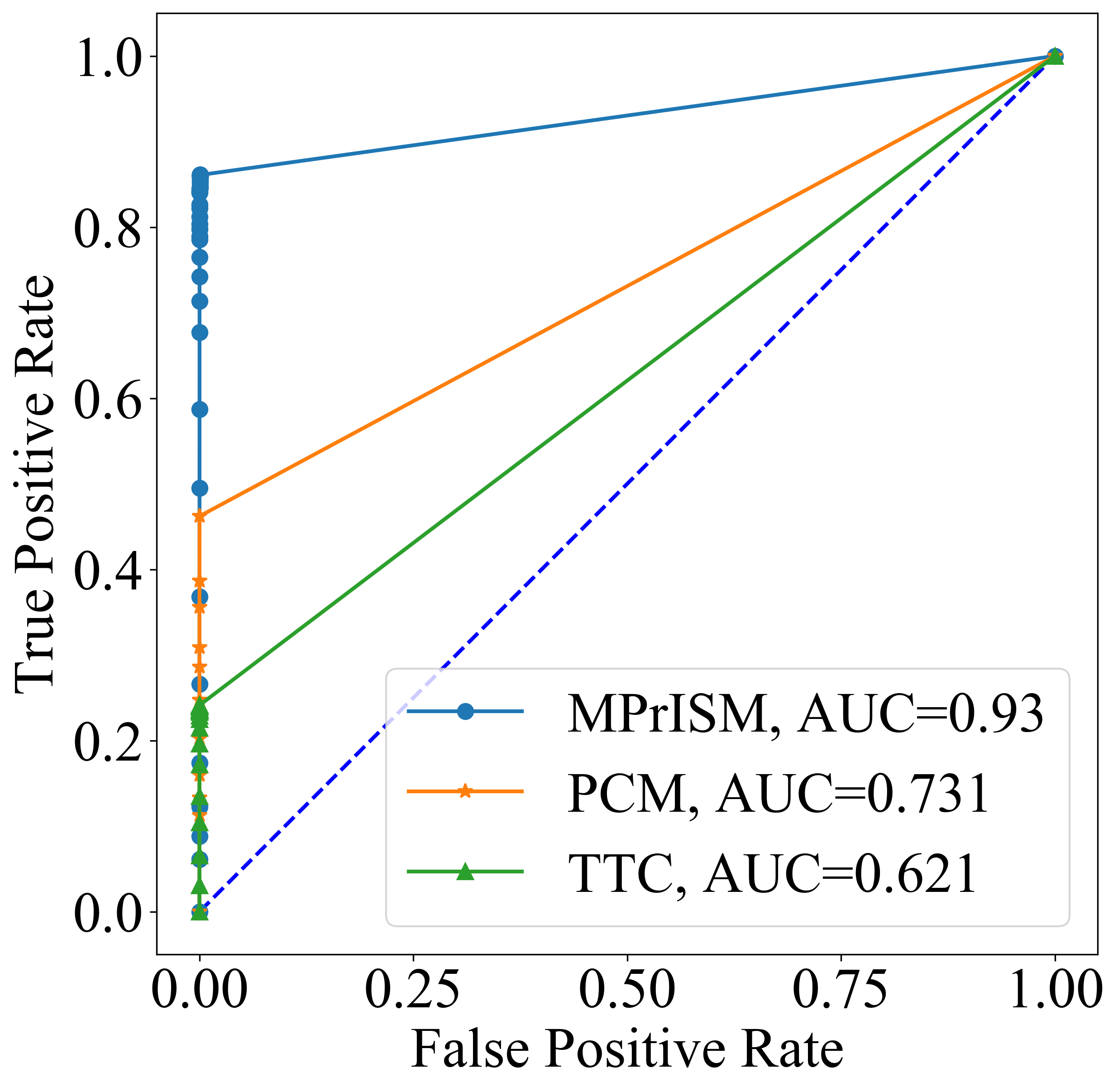} \\
            \includegraphics[width=1\textwidth]{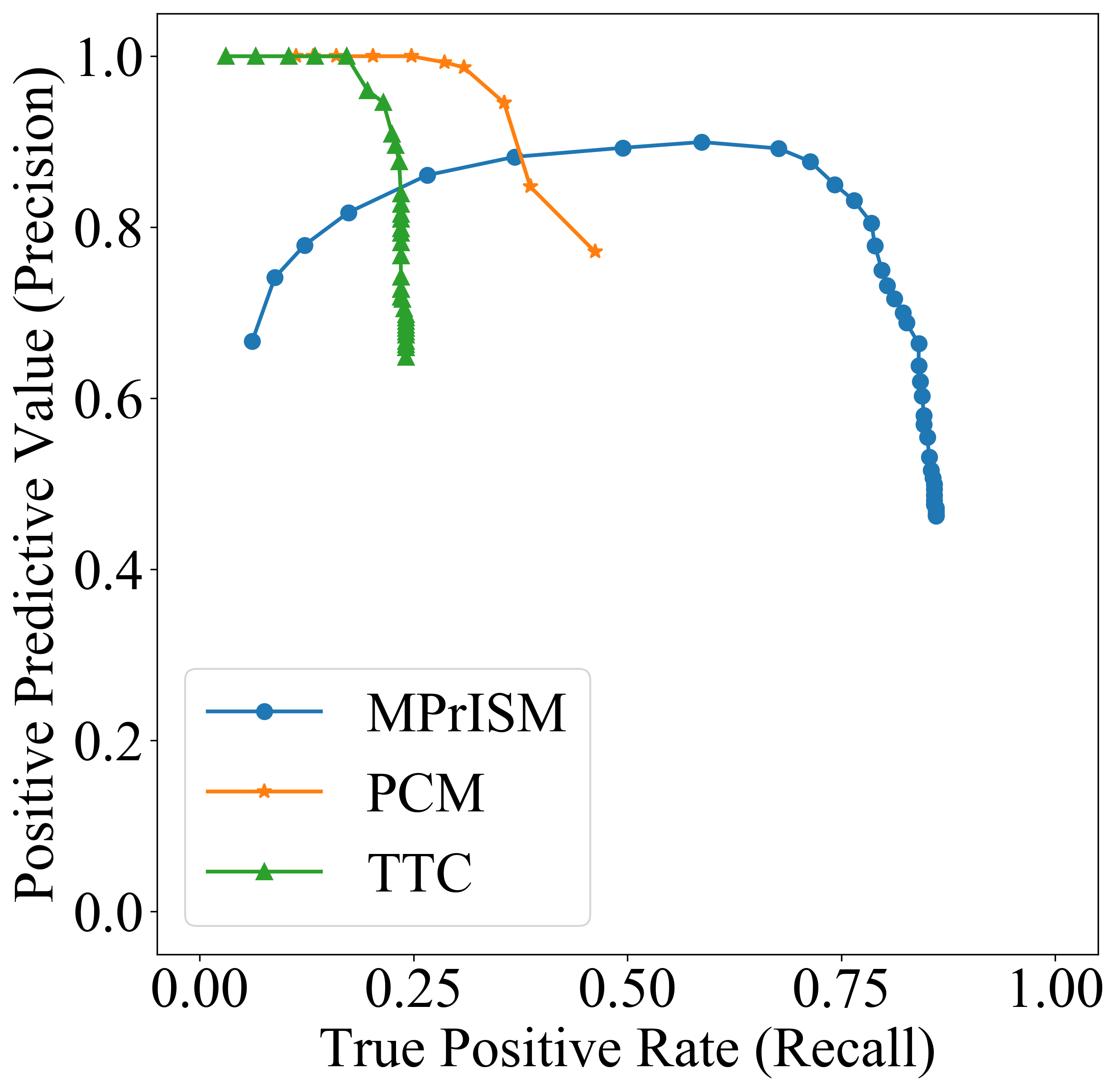}
            \end{minipage}
            }
        \subfloat[\label{fig:Safety-metrics-performance-varying-1.0}]{
            \begin{minipage}[b]{0.32\textwidth}
            \includegraphics[width=1\textwidth]{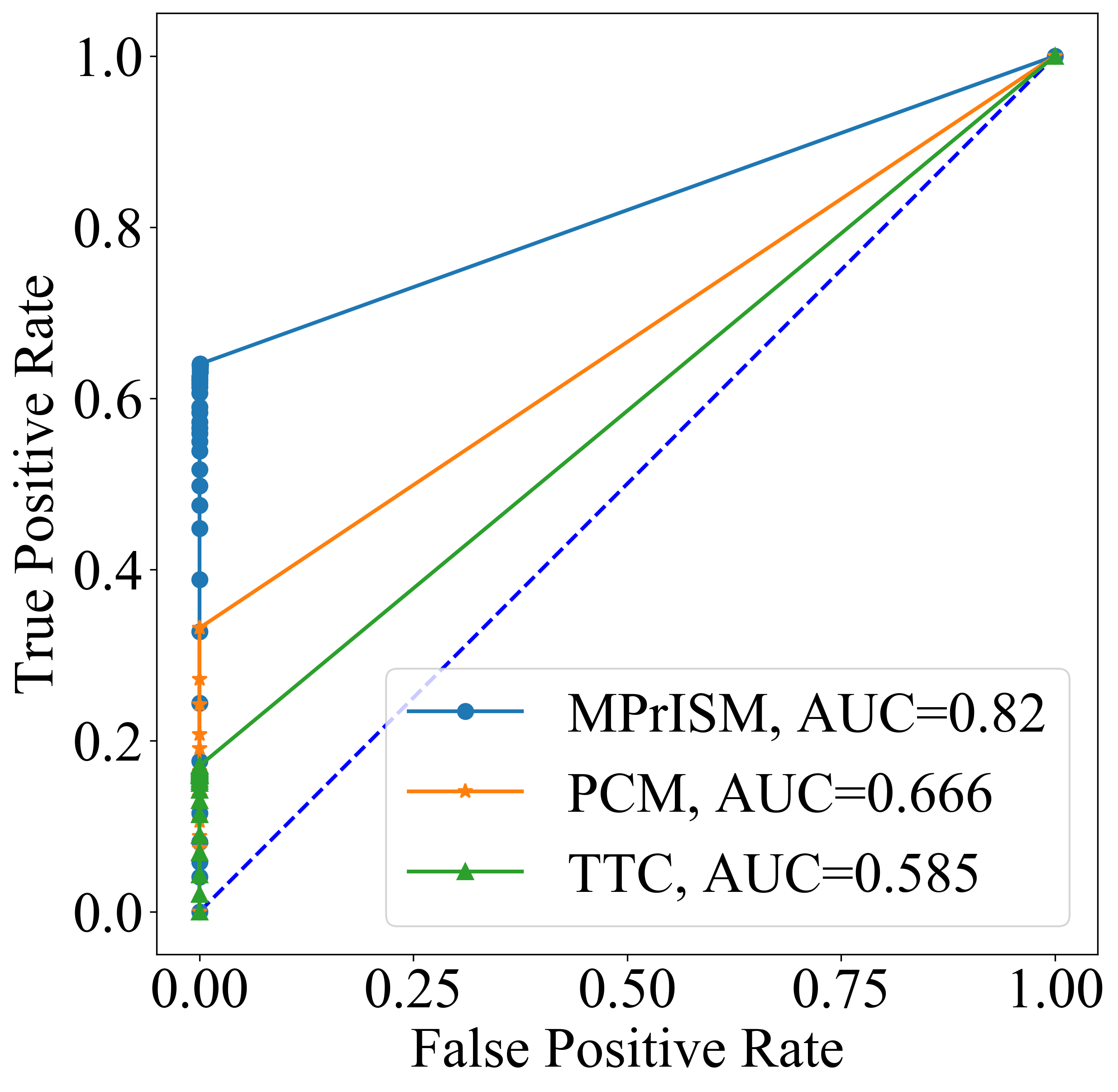} \\
            \includegraphics[width=1\textwidth]{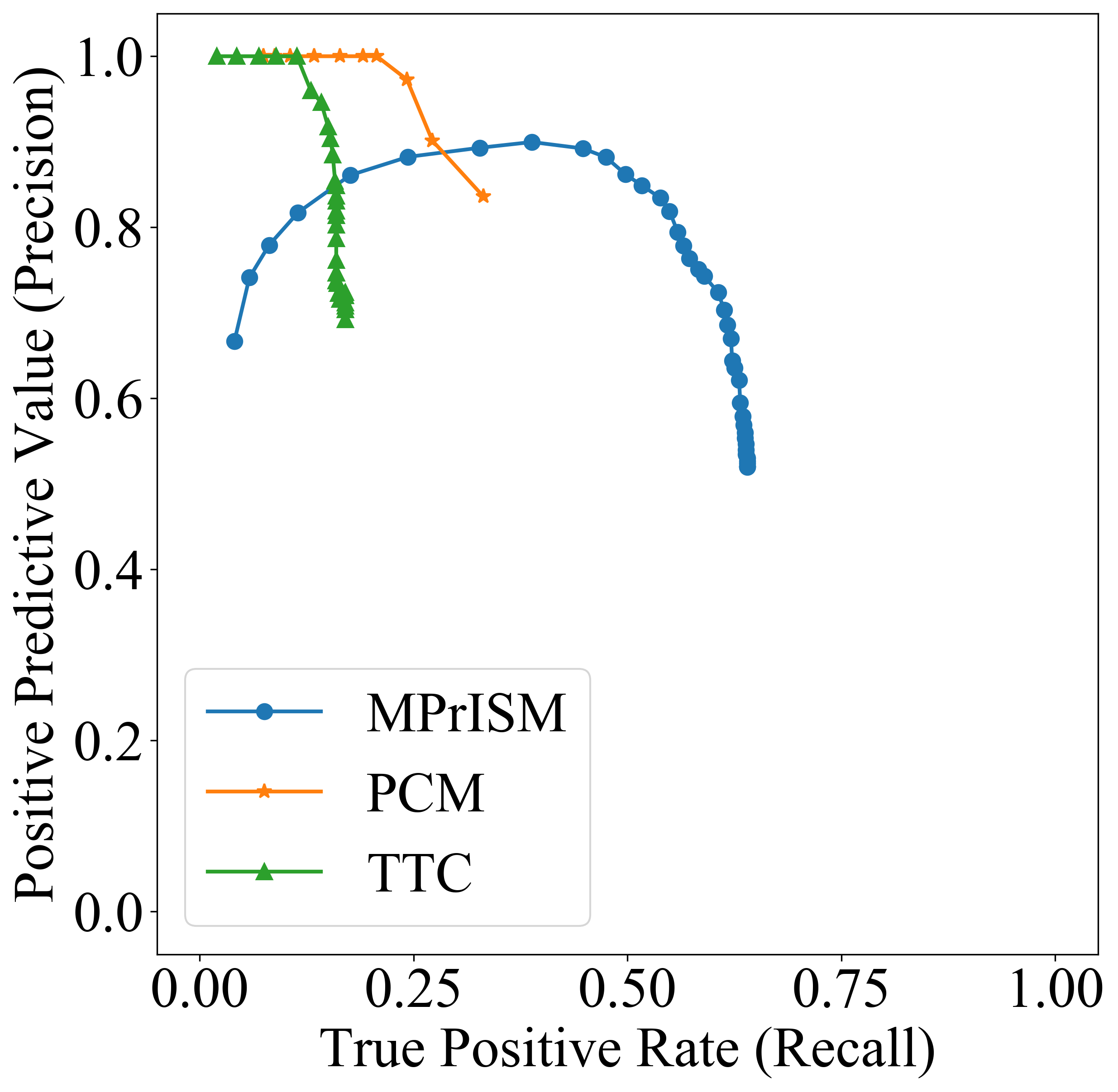}
            \end{minipage}
            }        
        \caption{ROC curve and PR curve for alarm (a)
        $0s$, (b) $0.5s$, and (c) $1.0s$ before the collision unavoidable moment. The first and second rows are the ROC and PR curves under different settings, respectively.\label{fig:Safety-metrics-performance-varying}}
\end{figure*}

From logged vehicle trajectory calculation results, we can find that the moment
110 is not dangerous and only MPrISM alarms at the moment. The MPrISM
predicted trajectories of SV and POV are shown in Fig. \ref{fig:Scenario4-MPrISM-traj}.
Since MPrISM assumes the BV will take the worst-case behaviors to
create safety-critical situations, it predicts the BV will continue
to steer into the SV lane. However, in the real trajectory, the BV
only makes lane change to the SV\textquoteright s adjacent lane. As
a result, the MPrISM produces false-alarm cases in this scenario.

\subsection{Statistical performance analysis of real-time safety metrics}

In this subsection, we analyze the statistical performance of real-time
safety metrics. To evaluate the predictive capability of real-time
safety metrics, different alarm times before the collision unavoidable moment, including $0s,0.5s,$ and $1.0s$, are examined and results are shown in Fig. \ref{fig:Safety-metrics-performance-varying}. Safety metrics should alarm in advance to indicate the driver or AV decision module. 

The results of the ROC curve and PR curve for the alarms at
collision unavoidable moments are shown in Fig. \ref{fig:Safety-metrics-performance-varying-0}.
Note again that each point in the figure corresponds to a specific
threshold parameter for each safety metric. The point (0,0) and (1,1)
in the ROC curve are situations that the safety metric never alarms
or alarms at all moments, correspondingly. From the ROC curve results, we can find that the MPrISM has the largest
AUC and TTC has the smallest AUC. Therefore, MPrISM has the best performance
in the ROC measurement comparing with the PCM and TTC on the tested
trajectory dataset. Note that since TTC can only detect longitudinal
dangerousness brought by the leading vehicle, it may fail in the cases
where BV collides the AV from behind as well as some angle
and sideswipe cases. For the PCM, since it neglects vehicles behind
the SV \cite{junietz2018criticality-PCM}, it may fail in the cases
where the BV collides with the AV from behind. 

From the PR curve results (lower row, Fig. \ref{fig:Safety-metrics-performance-varying-0}), the MPrISM can achieve the highest recall
but its precision is generally lower than the PCM and TTC. The PCM
generally performs better than the TTC since it can always achieves higher recall while achieving similar precision. As the MPrISM uses the worst-case behavior
assumption, it can capture most of the dangerous situations and therefore
has the highest recall. However, it may also introduce more false
alarms as demonstrated in the failure analysis. The PCM uses the constant
behavior assumption so it will produce fewer false alarm cases (higher
precision) and more false-negative cases (lower recall). As discussed
previously, there usually exists a trade-off between precision and
recall. Therefore, different safety metrics may be preferred depending
on the applications. For example, if the safety metric is used as
the safety guard for an AV decision module, then the recall is more important
since it is desired to capture all potentially dangerous situations.
In that case, MPrISM might be more suitable and can achieve better
performance since it can achieve the highest recall with reasonable
precision. On the contrary, if the safety metric is used for the on-road
test to evaluate a vehicle's safety performance, then precision is
more important. In this case, the PCM might have better performance
comparing with TTC and MPrISM. 

The ROC and PR curve results for alarms at $0.5s$ and $1.0s$ before the collision unavoidable moment are shown in Fig. \ref{fig:Safety-metrics-performance-varying-0.5}
and Fig. \ref{fig:Safety-metrics-performance-varying-1.0}, respectively.
The trend of the ROC and PR curves of each safety metric is similar
to that shown in Fig. \ref{fig:Safety-metrics-performance-varying-0} and generally shift toward the upper
left. With an earlier required alarm time, the prediction power for the safety metrics decreases, therefore the recall generally decreases. As shown in the results, the AUC decreases for all three
metrics. The precision of TTC and PCM do not change significantly because, in general, an earlier alarm time will not impact significantly on the number of their predicted positives. We should further note that the precision of MPrISM is higher for alarms before the collision unavoidable moment (Fig. \ref{fig:Safety-metrics-performance-varying-0.5} and Fig. \ref{fig:Safety-metrics-performance-varying-1.0}) than those alarms at the collision unavoidable moment (Fig. \ref{fig:Safety-metrics-performance-varying-0}). This is due to the worst-case behavior assumptions made in MPrISM, which can help capture dangerous situations that lead to the crash. 

\begin{figure}
\begin{centering}
\includegraphics[width=0.8\columnwidth]{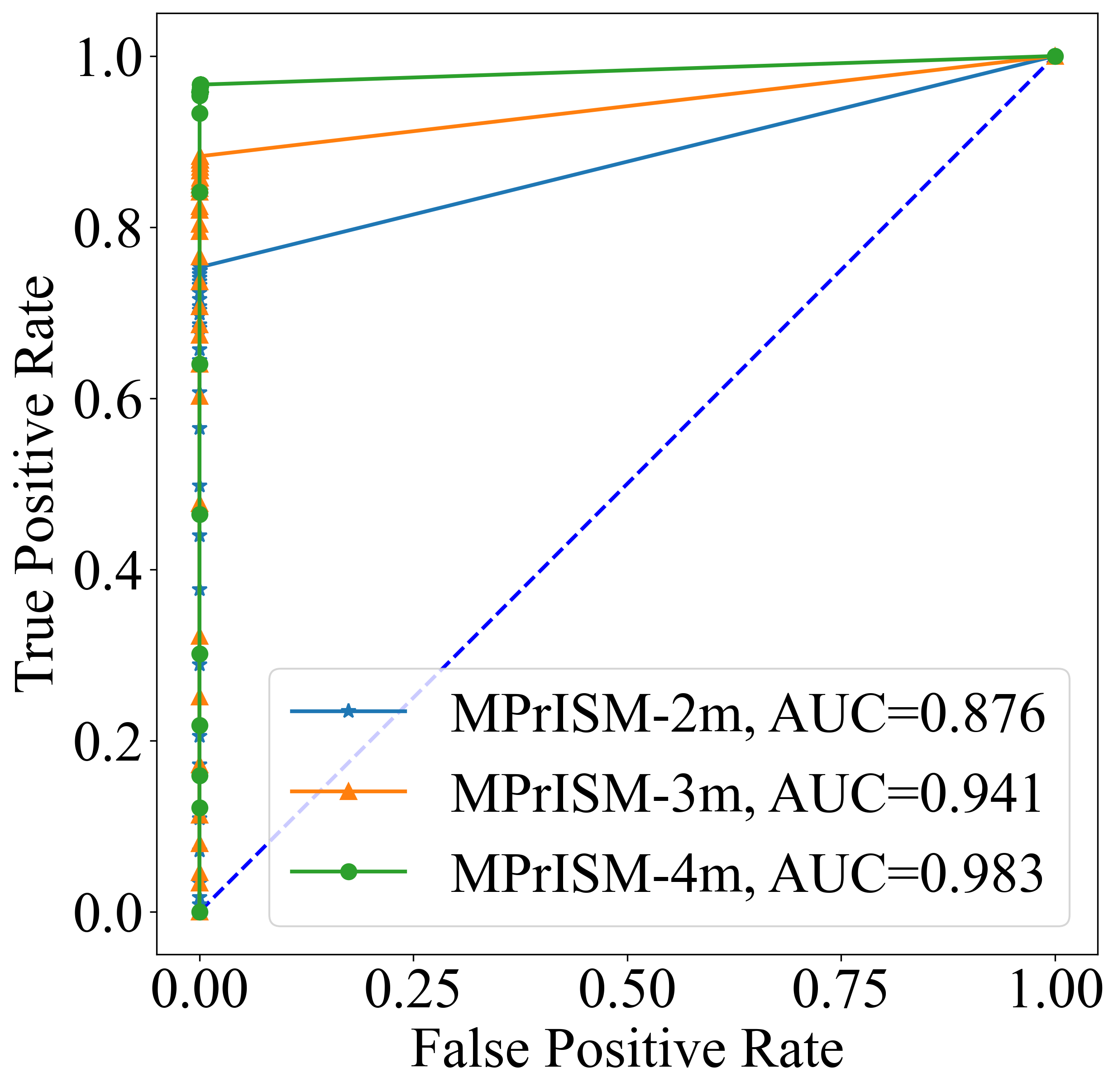}
\includegraphics[width=0.8\columnwidth]{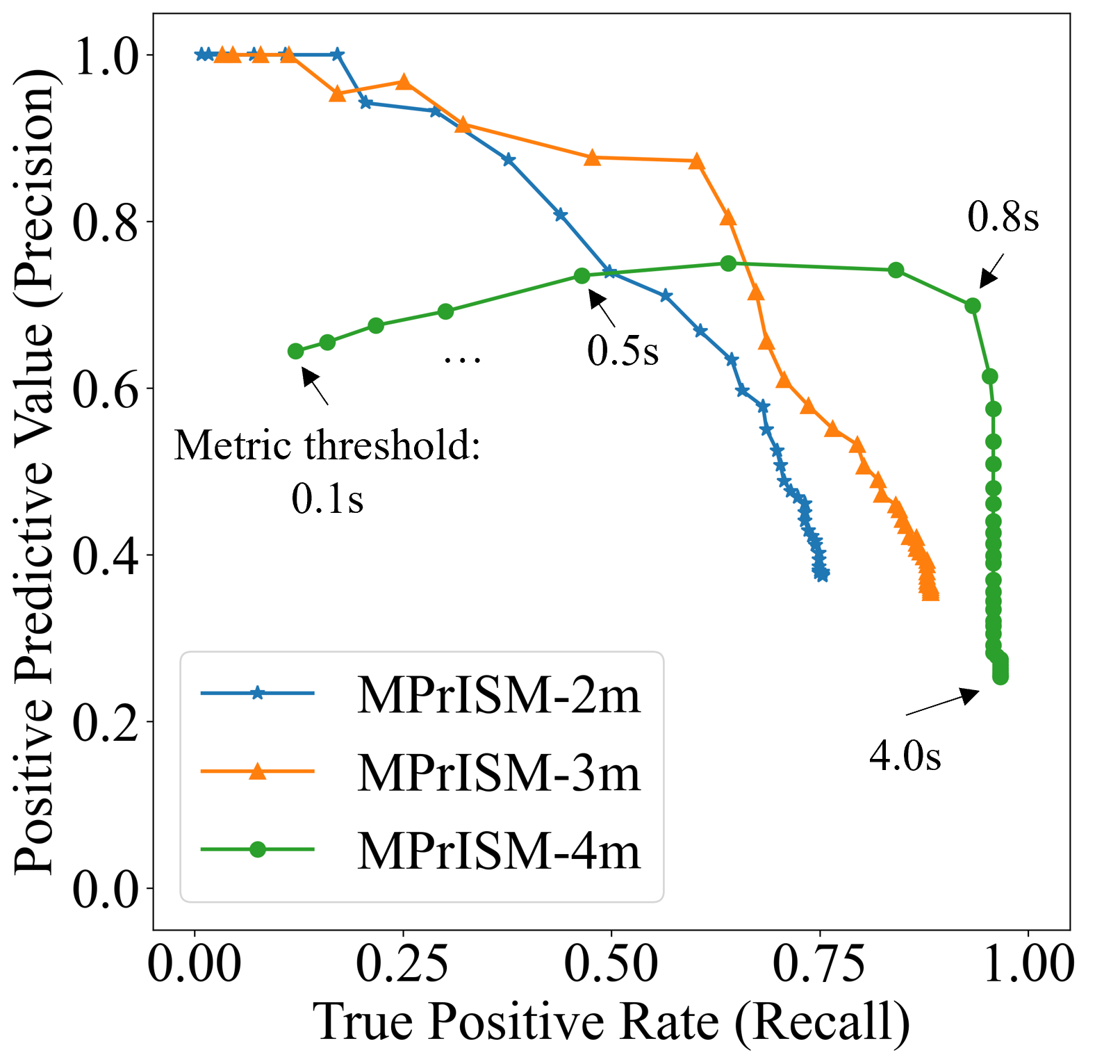}
\par\end{centering}
\caption{ROC curve and PR curve for MPrISM with different hyper-parameters and metric thresholds.} \label{fig:fine-tune}
\end{figure}


To further demonstrate the capability of the proposed method, we investigate how tuning hyper-parameters and metric activation thresholds could affect the ROC and PR curves, which is beneficial for model parameter selection. We use MPrISM as an example and examine the influence of collision threshold $C$. We set $C$ equal to 2, 3, and 4 meters, and the ROC and PR curves are shown in Fig. \ref{fig:fine-tune}.


From the ROC curve results, we can find that $C$=4m achieves the best AUC performance, since a larger collision threshold can capture more potentially dangerous situations. From the PR curve results, we can find that there is a trade-off between precision and recall. As we have discussed, a larger $C$ can generate a higher recall, but at the same time, it will also produce more false-positive alarms and lead to low precision.


Note that each data point on the ROC and PR curves corresponds with a different metric activation threshold. We choose the threshold ranges from 0.1s to 4.0s with a 0.1 step size. We only mark several metric activation thresholds in the PR curve as shown in Fig. \ref{fig:fine-tune} (bottom) for a clean visualization. From the PR curve results, we can see that a smaller metric activation threshold will lead to a higher precision but lower recall, and a higher threshold will produce better recall but worse precision. In general, we prefer the threshold that is close to (1,1) point on the PR curve, which balances the trade-off between precision and recall. Therefore, 0.8s seems to work the best with the testing dataset and these results demonstrate the capability of the proposed method for metric thresholds selection.


\subsection{Demonstration of proposed method with real-world trajectory data}

\begin{figure*}
\begin{centering}
\includegraphics[width=0.95\textwidth]{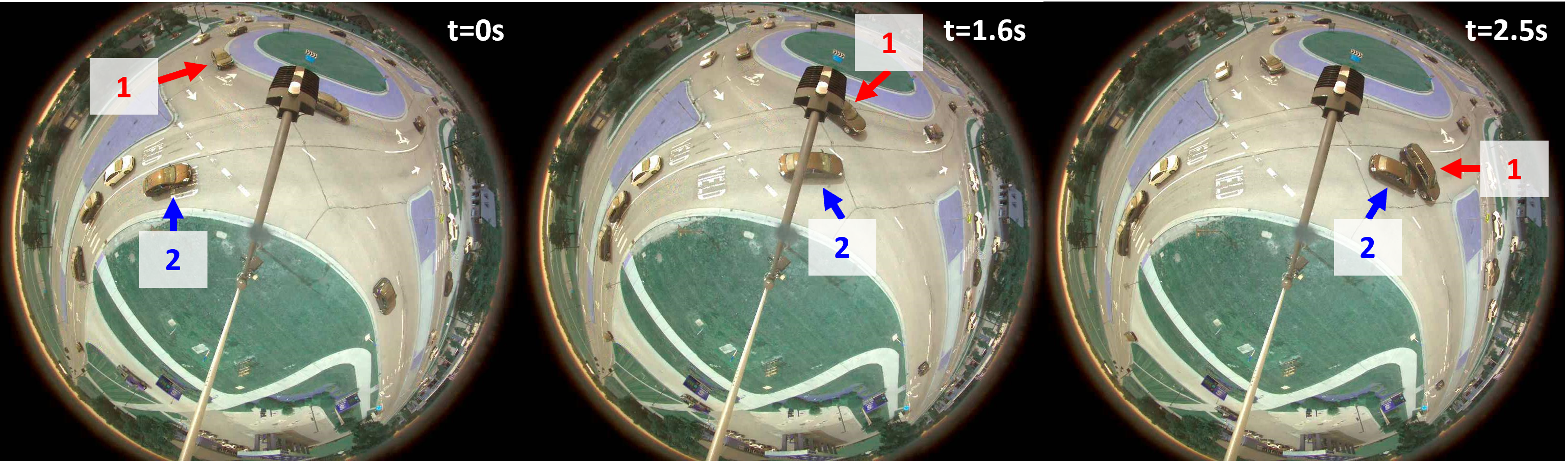}
\par\end{centering}
\caption{Real-world crash event at a two-lane roundabout located at Ann Arbor, Michigan.} \label{fig:real-world-crash}
\end{figure*}


We use real-world trajectory data, to further demonstrate that the proposed method is applicable to real-world dataset and diverse scenarios. The data is collected from a two-lane roundabout located at State St. and W Ellsworth Rd. intersection, Ann Arbor, Michigan, USA. A roadside perception system \cite{zou2022real, zhang2022design} is developed to collect vehicle trajectory data at 10Hz. We used real-world crash event data as demonstrated in Fig. \ref{fig:real-world-crash}. In this scenario, vehicle \#1 (shown in red) is circulating within the roundabout, and vehicle \#2 (shown in blue) is at the roundabout entrance. As captured by the roadside camera, vehicle \#2 fails to yield to the right-of-way of vehicle \#1, and a collision happened. 

\begin{figure}
\begin{centering}
\includegraphics[width=1.0\columnwidth]{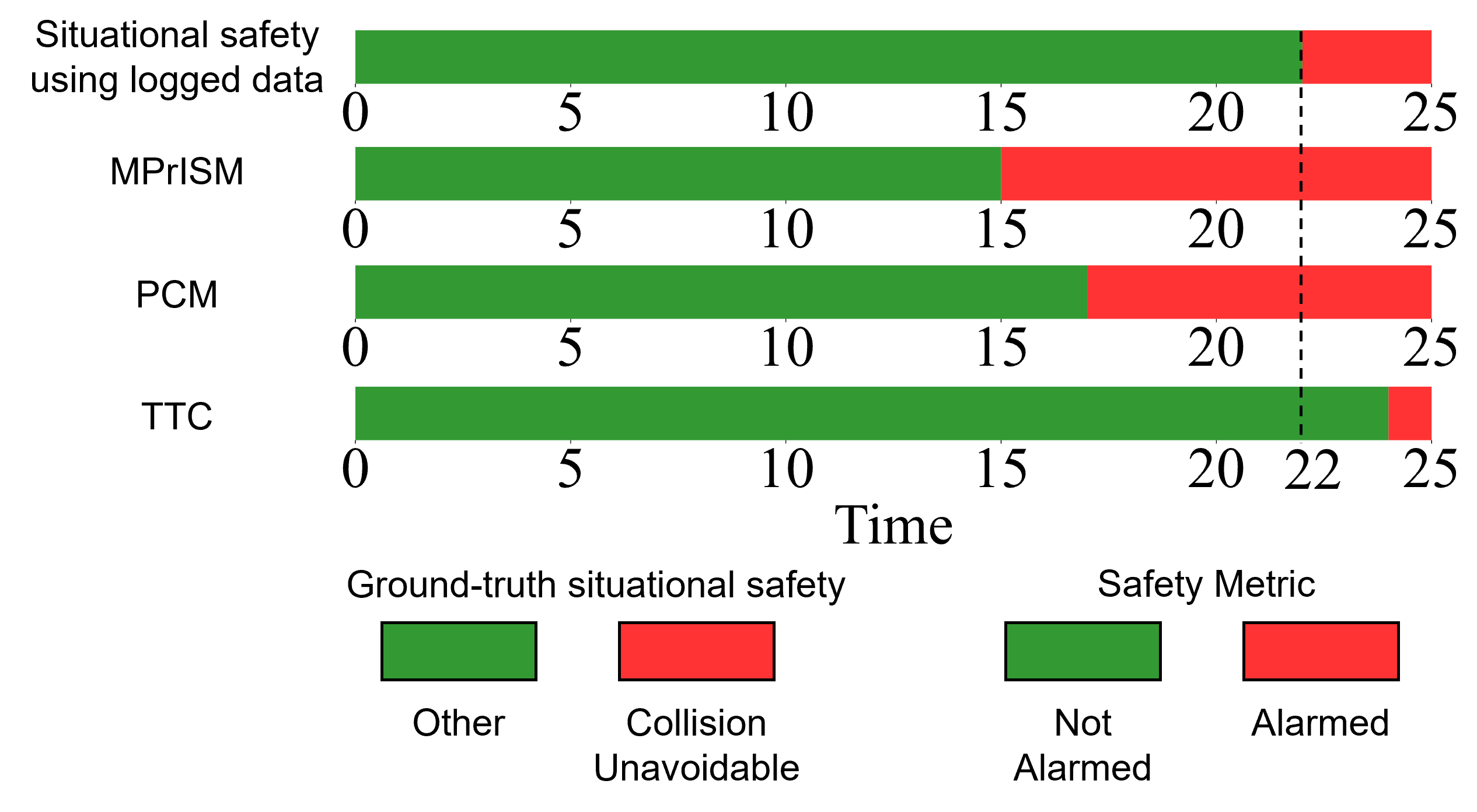}
\par\end{centering}
\caption{The situational safety calculated using logged trajectory data and safety metrics results of Ann Arbor crash data.\label{fig:AA-results}}
\end{figure}


We chose vehicle \#1 as the SV to evaluate its safety. The safety metrics results and the situational safety calculated using logged trajectory data are shown in Fig. \ref{fig:AA-results}. From logged vehicle trajectory calculation results, we can find that starting from moment 22, the SV is in a collision unavoidable situation. Both MPrISM and PCM can successfully alarm in this safety-critical event. They also indicate the dangerousness 0.7s and 0.5s before the collision unavoidable moment, respectively, which demonstrate their good predictive capabilities. However, TTC generates false-negative cases in this scenario since it fails to identify the conflicting vehicle as its leading vehicle. These results demonstrate that the proposed method can work with real-world data with good performance.

\section{Conclusion \label{sec:Sec-4}}

In this paper, we propose an evaluation framework to evaluate real-time safety metrics performance using logged vehicle trajectory data.
Specifically, an SV evasive trajectory planning problem
is formulated to determine whether there exists a safe trajectory
for the SV, given all near-future trajectories of surrounding BVs.
By analyzing the safety metric outputs and ground-truth situational safety
in each trajectory, we conduct failure analysis to identify the
metric failed cases and investigate potential weaknesses caused by
model assumptions, approximations, and parameters. Furthermore, by
accumulating large quantities of simulated trips, the statistical performance of real-time
safety metrics can be accurately evaluated. The proposed method is
used to evaluate three representative real-time safety metrics, including
TTC, PCM, and MPrISM. The results show that the method can successfully
identify potential weaknesses of all three metrics, which are valuable
for further refinements. Moreover, the results validate
the capability of the proposed method to analyze and characterize
the statistical performance of different metrics. Specifically, the
MPrISM can achieve the highest recall and the PCM has the best accuracy.
The evaluation and comparison between metrics are very important to
help researchers, practitioners, and regulators to choose from 
existing metrics for different application purposes. 

For future research, the simulated trajectory dataset used in this study can be further enriched
and improved. With a larger data size and broader coverage of scenarios,
the real-world driving environment can be better represented and therefore
safety metrics can be evaluated with more scenarios. With
the development of sensing technology and the data acquisition system,
real-world trajectories can be recorded and used to complement the
simulated trajectories.
Real-world data provides higher fidelity than simulation data, particularly in areas such as vehicle dynamics. This makes it valuable for accurately analyzing metrics performance and we leave that for future study.

\appendix[Safety metrics model parameters\label{appendix}]

For the TTC, a vehicle is considered as the ego vehicle's leading
vehicle if their lateral offset is smaller than a threshold ($2m$).
It alarms if the output TTC is smaller than $1$ second.

For the PCM, following the notation in \cite{junietz2018criticality-PCM},
the model parameters are as follow: weighting factors $w_{x}=1,w_{y}=1,w_{ax}=0.1,w_{ay}=1$
, time resolution 0.1$s$, look-ahead steps 20. According to the original
paper, both the criticality and maximum expected acceleration of the
moment can be used as the safety surrogate and we choose the latter
one. For the PCM, it alarms if the maximum expected acceleration equals
to or greater than a predefined threshold (8.0$m/s^{2}$). 

For the MPrISM, following the notation in \cite{weng2020-MPrISM},
the model parameters are as follows: time resolution $\Delta=0.1s$,
look-ahead steps $T=10$, collision threshold $C=4m$. For the MPrISM,
it alarms if the output model predictive time-to-collision (MPrTTC)
is smaller than $1$ second. 

For the ground-truth situational safety calculation using logged vehicle trajectory data, the model parameters are as follows:
look-ahead times 20, time resolution $0.1s$, radius $\delta=1.3m$, circle
center distance $l=3.5m$. 

\begin{figure}[h]
\begin{centering}
\includegraphics[width=0.9\columnwidth]{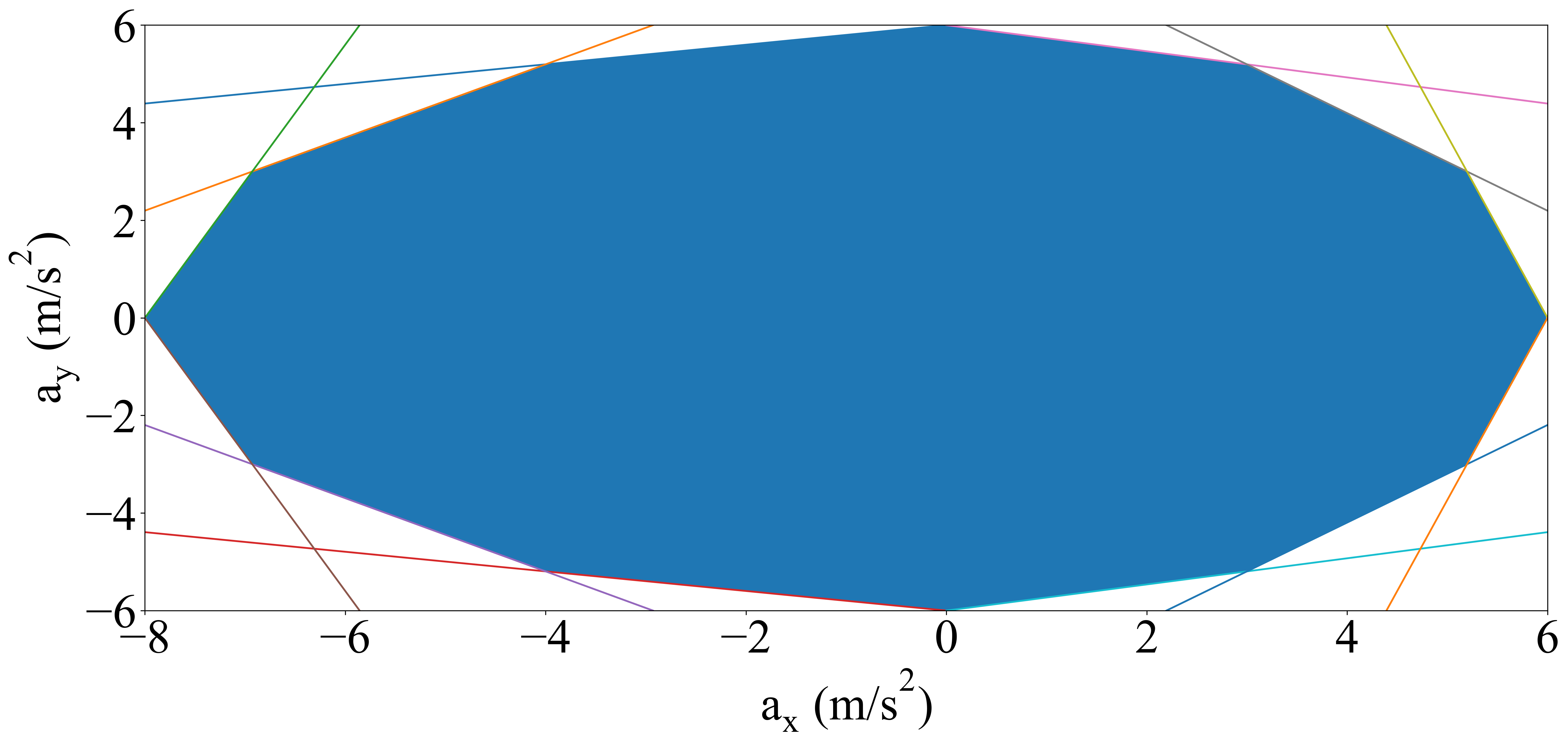}
\par\end{centering}
\caption{Kamm's circle. \label{fig:Kamm's-circle}}
\end{figure}

The Kamm\textquoteright s circle used in PCM, MPrISM, and the ground-truth situational safety calculation is shown in Fig. \ref{fig:Kamm's-circle}.

%



\section*{Acknowledgment}
The views presented in this paper are those of the authors alone. The authors would like to thank Dr. Bowen Weng and Dr. Philipp Junietz for their help in implementing the MPrISM and PCM algorithms.

\ifCLASSOPTIONcaptionsoff
  \newpage
\fi



\bibliographystyle{IEEEtran}
\bibliography{./bibtex/bib/IEEEabrv,./bibtex/bib/reference}

%



%


\vfill
\begin{IEEEbiography}[{\includegraphics[width=0.9in,height=1.25in,clip,keepaspectratio]{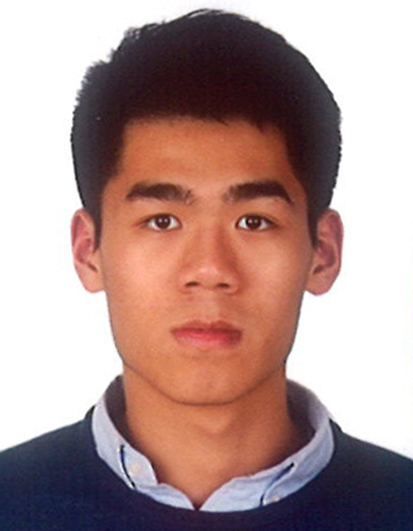}}]{Xintao Yan}
    received the bachelor's degree in automotive engineering from
    Tsinghua University in 2018 and he is currently pursuing his Ph.D. degree in the Department of Civil and Environmental Engineering at the University of Michigan, Ann Arbor. His research interests are mainly about the safety of connected and automated vehicles, including naturalistic driving behavior modeling and automated driving system evaluation. He was the recipient of the Exceptional Paper Award from the Transportation Research Board (TRB) Annual Meeting in 2019 and the Intelligent Transportation Systems Best Paper Award from the INFORMS Transportation Science and Logistics society in 2021. 

\end{IEEEbiography}

\begin{IEEEbiography}[{\includegraphics[width=1.30in,height=1.40in,clip,keepaspectratio]{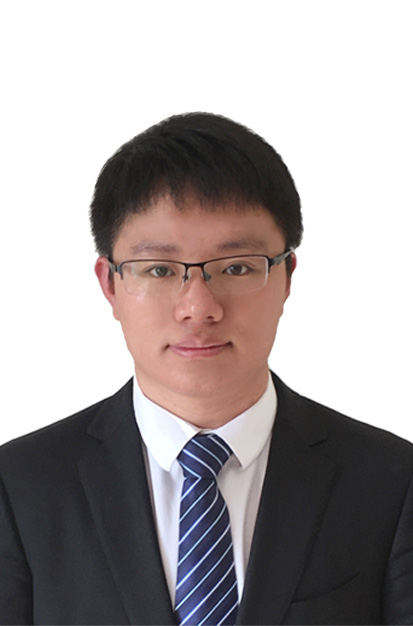}}]{Shuo Feng}
	(Member, IEEE) received the bachelor's and Ph.D. degrees in the Department of Automation at Tsinghua University, China, in 2014 and 2019, respectively. He was a postdoctoral research fellow in the Department of Civil and Environmental Engineering and also an Assistant Research Scientist at the University of Michigan Transportation Research Institute (UMTRI) at the University of Michigan, Ann Arbor. He is currently an Assistant Professor in the Department of Automation at Tsinghua University. His research interests lie in the development and validation of safety-critical machine learning, particularly for connected and automated vehicles. He is the Associate Editor of the IEEE Transactions on Intelligent Vehicles and the Academic Editor of the Automotive Innovation. He was the recipient of the Best Ph.D. Dissertation Award from the IEEE Intelligent Transportation Systems Society in 2020 and the ITS Best Paper Award from the INFORMS TSL society in 2021.
\end{IEEEbiography}

\begin{IEEEbiography}[{\includegraphics[width=1.15in,height=1.2in,clip,keepaspectratio]{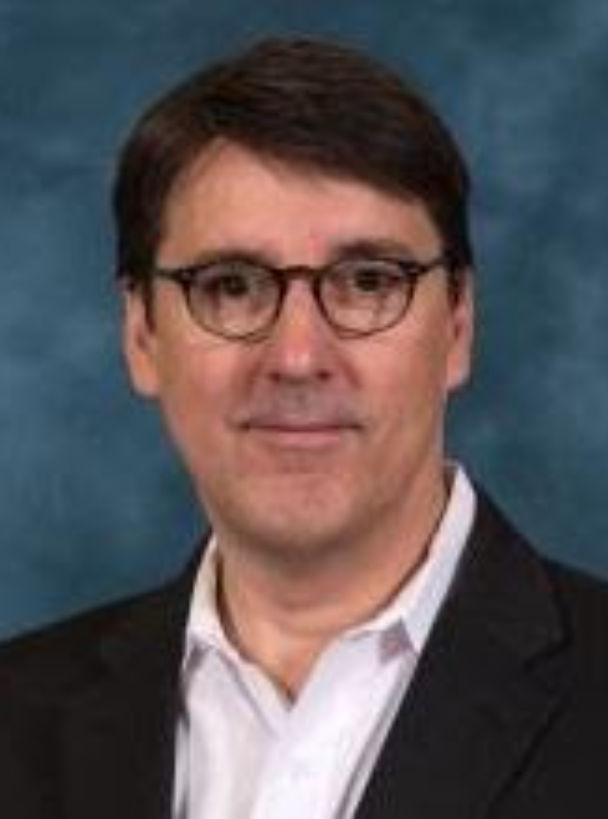}}]{David J. LeBlanc} 
    received the      bachelor’s and
    master’s degrees in mechanical engineering from
    Purdue University and the Ph.D. degree in aerospace
    engineering from University of Michigan. He has
    been with University of Michigan Transportation
    Research Institute since 1999, where he is currently
    an Associate Research Scientist. His work focuses
    on the automatic and human control of motor vehicles, particularly the design and evaluation of driver
    assistance systems.
    
\end{IEEEbiography}

\begin{IEEEbiography}[{\includegraphics[width=1.15in,height=1.2in,clip,keepaspectratio]{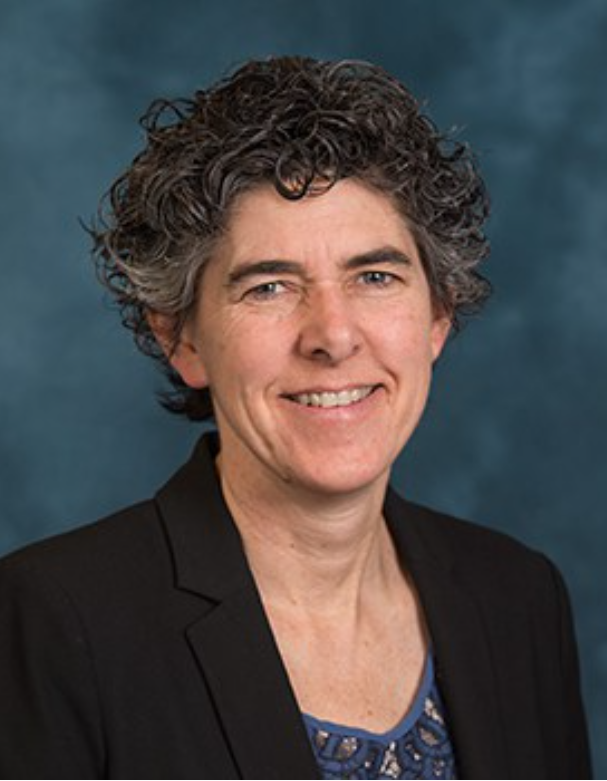}}]{Carol Flannagan}
    is a Research Professor in UMTRI's Biosciences Group, and Director of the Center for the Management of Information for Safe and Sustainable Transportation CMISST, at UMTRI. She serves on the Statistics and Methods section of the Research Core at the U-M Injury Prevention Center (IPC) providing support for transportation methods, data, and analysis. Dr. Flannagan is also the Motor Vehicle Crash (MVC) Content Lead for the IPC Center, providing expertise in the area of MVC prevention science. Dr. Flannagan has over 20 years of experience conducting data analysis and research on injury risk related to motor vehicle crashes and was responsible for the development of a model of injury outcome that allows side-by-side comparison of public health, vehicle, roadway and post-crash interventions.

\end{IEEEbiography}

\begin{IEEEbiography}[{\includegraphics[width=1.15in,height=1.2in,clip,keepaspectratio]{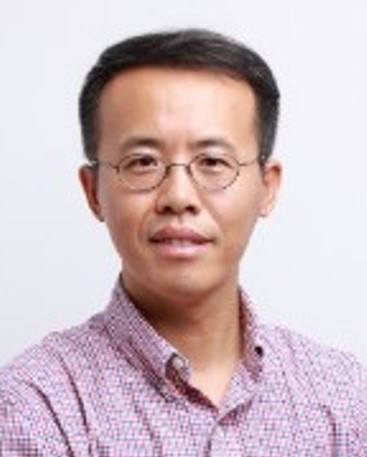}}]{Henry X. Liu}
    (Member, IEEE) received the
    bachelor's degree in automotive engineering from
    Tsinghua University, China, in 1993, and the Ph.D.
    degree in civil and environment engineering from
    the University of Wisconsin-Madison in 2000. He is currently the Director of Mcity and a Professor of Civil and Environmental Engineering at the University of Michigan, Ann Arbor. He is also the Director for the Center for Connected and Automated Transportation (USDOT Region 5 University Transportation Center) and a Research Professor at the University of Michigan Transportation Research Institute. Prof. Liu conducts interdisciplinary research at the interface of transportation engineering, automotive engineering, and artificial intelligence. Specifically, his scholarly interests concern traffic flow monitoring, modeling, and control, as well as training and testing of connected and automated vehicles. He has published more than 130 refereed journal papers and his work have been widely recognized in the public media for promoting smart transportation innovations. Prof. Liu is also the managing editor of Journal of Intelligent Transportation Systems.
\end{IEEEbiography}





\end{document}